\PassOptionsToPackage{table}{xcolor}
\documentclass[11pt]{article}

\usepackage[final]{acl}

\usepackage{times}
\usepackage{latexsym}

\usepackage[T1]{fontenc}

\usepackage[utf8]{inputenc}

\usepackage{microtype}

\usepackage{inconsolata}

\usepackage{graphicx}

\usepackage{amsmath}
\usepackage{amssymb}
\usepackage{mathtools}
\usepackage{amsthm}

\usepackage[table]{xcolor}

\usepackage{subcaption}
\usepackage{booktabs} 
\usepackage{multirow}
\usepackage{tcolorbox}

\definecolor{lightgreen}{RGB}{240, 255, 240}
\definecolor{lightred}{RGB}{255, 245, 245}
\definecolor{lightblue}{RGB}{240, 248, 255}
\definecolor{lightyellow}{RGB}{255, 250, 240}

\usepackage[capitalize,noabbrev]{cleveref}

\theoremstyle{plain}

\theoremstyle{definition}

\theoremstyle{remark}

\usepackage[textsize=tiny]{todonotes}

\title{Diverse Thinking Schemata Elicit Better Reasoning\\ in Large Language Models}

\author{Xinyue Liang, Yizhe Yang, Yu Bai, Bin Xu, Jiawei Li, Yang Gao\thanks{Corresponding author.} \\
       School of Computer Science and Technology,\\
Beijing Institute of Technology, Beijing, China\\
\texttt{\{xyliang, gyang\}@bit.edu.cn} \\}

\begin{document}
\maketitle
\begin{abstract}
  Large reasoning models (LRMs) have attracted increasing attention for their ability to solve complex mathematical problems by generating extended reasoning chains. 
  In this work, we focus on two critical yet underexplored aspects of the reasoning process: \textit{reasoning transitions} capturing the distinct transitions between reasoning steps and \textit{answer candidates} reflecting the variety of solution paths produced by the model. We collectively define these two aspects as \textbf{thinking schemata}.
  We observe a correlation between the diversity of thinking schemata and model performance, which motivates us to enhance diversity as a means to further improve reasoning potential. To this end, we propose \textbf{Di}verse \textbf{Sc}hemata Policy \textbf{O}ptimization (DiScO), a framework that first endows the model with schemata awareness, then encourages diversity through reinforcement learning, and further promotes diverse reasoning at inference time. Experiments on multiple mathematical reasoning benchmarks demonstrate that DiScO consistently outperforms standard group relative policy optimization. 
  Beyond accuracy, human-annotated analyses show that DiScO substantially improves the model's ability to recover from erroneous initial attempts. 
  Overall, our work suggests the important role that diversity of the thinking schemata plays and points to scaling along the diversity dimension as a promising research direction.
\end{abstract}

\section{Introduction}

Large reasoning models (LRMs) have emerged as a promising paradigm for complex problem solving, where performance is achieved through generating multi-step chains of thought that decompose tasks into intermediate reasoning steps. Recent advances in scaling and reinforcement learning have further strengthened their ability to tackle challenging reasoning tasks~\citep{deepseekai2025deepseekr1}. Nevertheless, current approaches often impose overly predefined reasoning structures, such as strictly chain-like sequences~\citep{wei2022chain, zhangautomatic} or explicitly organized trees and graphs~\citep{yao2023tree, besta2024graph, yao2024got}.

\begin{figure*}[t]
\centering
\includegraphics[width=\linewidth]{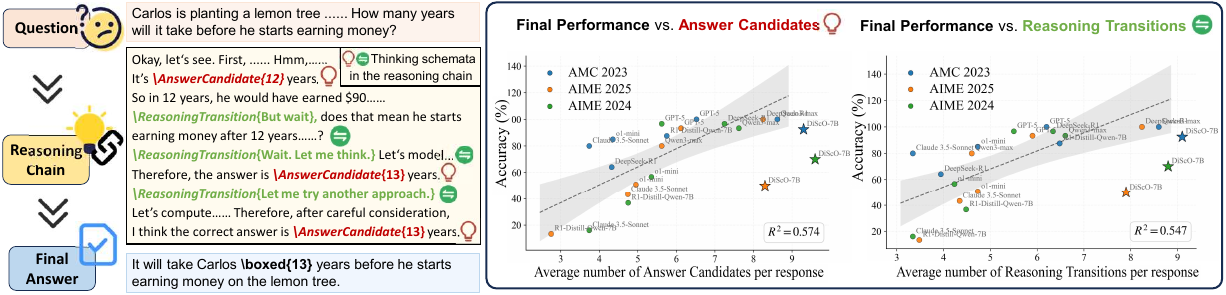} 
\caption{An illustration of the thinking schemata, including answer candidates and the reasoning transitions, that we defined and their relation with the final performance. An enlarged  scatter plot is provided in Figure~\ref{fig:diversity-scatter-enlarged}.}
\label{fig:intro}
\end{figure*}

To better capture this cognitive flexibility, we introduce the concept of \textit{thinking schemata}, the latent reasoning structure that governs how the model transitions between intermediate thoughts and arrives at potential solutions during multi-step problem solving, inspired by schema theory in cognitive neuroscience~\citep{axelrod1973schema,arbib1992schema, fischbein1997schemata}. \footnote{Compared to the previous parlance of schemata~\citep{agarwal2025think, wen2025schema, chen2025schema, shen2025law} which mostly focus on the data with graph structure, the thinking schemata we defined in this work consider more regarding the thinking structure.} To make this concept analytically tractable, we define thinking schemata through two components that can be identified from a trajectory: the transitions (e.g., ``alternatively'', ``on the other hand'', etc.) between intermediate reasoning steps (i.e., \textit{Reasoning Transitions}) and the candidate solutions that emerge from various attempts (i.e., \textit{Answer Candidates}), shown in Figure~\ref{fig:intro}. Building on this formulation, we empirically investigate how thinking schemata relate to model performance. Our preliminary experiments reveal that the diversity of thinking schemata, characterized by the number of reasoning transitions and answer candidates, plays a crucial role in successful problem solving (see Section~\ref{sec:verification}). 

This observation motivates us to propose DiScO (\textbf{Di}verse \textbf{Sc}hemata Policy \textbf{O}ptimization), a framework designed to enhance the diversity of thinking schemata during both training and inference. During training, we first equip the model with basic self-awareness of its reasoning process by having it annotate points of thought transition and enumerate possible intermediate answers. We then apply reinforcement learning to encourage the generation of diverse transitions and distinct solution paths, thereby strengthening the model's reasoning capacity. At inference time, we further promote diversity through (i) truncating the initial portion of the reasoning chain to allow the model to restart from a fresh perspective, and (ii) filtering out duplicated reasoning to facilitate continued exploration when the maximum length is reached.

We evaluate DiScO on a range of challenging mathematical reasoning benchmarks and out-of-domain reasoning tasks, using both same-scale open-source models and frontier LLMs as reference points. Across the 7B and 32B scales, DiScO consistently improves over standard GRPO and achieves strong performance among open-source reasoning models. In addition to pass@1 gains, DiScO also improves pass@k performance, suggesting a stronger ability to explore diverse solution paths across multiple attempts. More importantly, human-annotated analyses show that models trained with DiScO recover more effectively from erroneous initial attempts and are less likely to discard correct intermediate answers. These results suggest that promoting diverse thinking schemata strengthens not only final-answer accuracy but also the model’s ability to self-correct during reasoning.

In summary, the key contributions are as follows:
\noindent
\begin{itemize}
\item We introduce the concept of thinking schemata, characterizing the latent reasoning structure through reasoning transitions and answer candidates, and empirically demonstrate that the diversity of thinking schemata correlates with reasoning performance.
\item We propose Diverse Schemata Policy Optimization (DiScO), a framework that enhances the diversity of thinking schemata during both training and inference.
\item Experimental results across five mathematical reasoning benchmarks and three out-of-domain benchmarks consistently demonstrate the effectiveness and generalization ability of DiScO, while also offering practical insights into the thinking process of LRMs.
\end{itemize}

\section{Related Works}

\paragraph{Large Reasoning Models} Recent advances in Large Reasoning Models (LRMs), such as OpenAI o1~\citep{jaech2024openai-o1} and DeepSeek-R1~\citep{guo2025deepseek}, highlight the role of reinforcement learning (RL) in enabling reasoning capabilities, with notable gains in mathematical reasoning~\citep{deepmath103k2025} and code generation~\citep{zhuo2024bigcodebench}. As RL becomes standard for improving LRMs, recent work has increasingly emphasized not only correctness but also diversity in reasoning~\citep{yao2025diversity}. Several methods focus on improving GRPO via more effective reward design, including length-dependent rewards and difficulty-aware reweighting~\citep{zhang2025grpo}, as well as adaptive shaping rewards via bi-level optimization~\citep{hu2020learning}. Given GRPO's widespread effectiveness, we also build on this framework.

\paragraph{Reasoning Patterns in LRMs} Researchers have investigated reasoning patterns in LRMs and their effects on problem-solving. \citet{minegishi2025topology} introduces graph-theoretic analysis of hidden-state clusters during reasoning. \citet{tian2025think} presents multi-round test-time thinking without chain-of-thought access. Studies reveal that LRMs construct structured, multi-stage chains~\citep{marjanovic2025deepseek}, though overly long reasoning (i.e. "overthinking") can degrade performance~\citep{an2025don}. \citet{yang2025reasonflux} enhances LRMs with hierarchical thought templates. While smaller LLMs benefit from structured patterns like decomposition and self-critique, larger models perform best with simpler monologue-style reasoning~\citep{wen2025thinkpatterns}. \citet{lee2025cot} proposes automatically extracting and clustering diverse chain-of-thought strategies. Our definition of schemata is more general than these thinking patterns, focusing on answer candidates and transitions between reasoning trajectories rather than manual classification.

\paragraph{Diversity in LRMs} Recent studies have explored diversity in LRM training. \citet{yao2025diversity} demonstrate a strong correlation between solution diversity and performance, integrating a token-level diversity objective into training. \citet{wang2025highentropy} identify high-entropy tokens as key divergence points. However, token-level methods operate on localized signals, encouraging only micro-level diversity. Trajectory-level approaches using GFlowNets~\citep{hu2023amortizing, yu2025flow, nair2025flowoptions, younsi2025accurate} sample reasoning traces proportionally to reward but rely on structural assumptions such as DAG-based formulations that can constrain the reasoning space. In contrast, our method directly evaluates complete rollout trajectories, capturing how they shift perspectives and explore alternative solution avenues, leading to genuine semantic diversity that better reflects human-like problem-solving.

\section{Thinking Schemata}
\subsection{Definitions}

\textit{Schemata}, originally introduced in cognitive science \citep{casson1983schemata, rumelhart1984schemata, bartlett1995remembering}, refer to structured frameworks or mental templates that guide perception, understanding, and reasoning~\citep{fischbein1999intuitions, fischbein1997schemata, cheng1985pragmatic}. Inspired by this theory, we define \textbf{Thinking Schemata} in the context of LRMs as the latent reasoning structure that governs how the model transitions between intermediate thoughts and arrives at potential solutions during multi-step problem solving. Thinking schemata provide a higher-level abstraction of the reasoning dynamics within a model, encompassing how it shifts perspectives, explores the solution space, and generates candidate reasoning paths.

In multi-step problem solving, LRMs generate a reasoning trajectory consisting of a sequence of intermediate thoughts before arriving at a final answer. Within this trajectory, two observable phenomena naturally emerge. First, the model may shift between semantically distinct reasoning perspectives, such as moving from numerical computation to algebraic manipulation, or from geometric intuition to formal derivation. We refer to these shifts as \textbf{Reasoning Transitions}. Second, the model may propose multiple plausible solutions or sub-conclusions throughout the reasoning process, particularly when facing tasks with ambiguity or uncertainty. We refer to these as \textbf{Answer Candidates}. Together, reasoning transitions and answer candidates constitute the observable manifestations of thinking schemata, reflecting how a model explores the solution space and navigates between different reasoning paths. 

Building on these two components, we further define the \textbf{Diversity of Thinking Schemata} as the variation in both reasoning transitions and answer candidates within a reasoning trajectory. A model with diverse thinking schemata demonstrates greater cognitive flexibility in exploring alternative reasoning routes and covers a broader inferential space. An illustrative example is provided in Appendix~\ref{app:schemata-example}.

\subsection{Diverse Thinking Schemata Are Associated with Better Reasoning}
\label{sec:verification}

To understand how thinking schemata relate to reasoning performance, we conduct a preliminary study examining the correlation between the diversity of thinking schemata and model accuracy across different LRMs.

We sample reasoning chains from a range of models, including GPT-5~\citep{singh2025openai}, DeepSeek-R1~\citep{deepseekai2025deepseekr1}, Qwen3, etc., on the three benchmarks~\citep{aops2023amc10a}. Each sampled chain is annotated by human annotators following the annotation protocol described in Appendix~\ref{app:human-study}, producing per-instance metrics including the number of Answer Candidates (``answerCandidates-avg'') and the number of semantic transitions (``reasoningTransition-avg'').


We quantify the relationship between model accuracy and diversity metrics by fitting linear regression models across multiple model-dataset points. As shown in Figure~\ref{fig:intro}, we observe a positive association between model accuracy and both diversity metrics. Notably, this correlation is not merely a byproduct of model capability, since stronger models like Claude 3.5-Sonnet do not consistently outperform the weaker DeepSeek-R1-Distill-Qwen-7B in diversity or accuracy on AMC 2023. This empirical finding suggests that LRMs benefit from exploring varied inferential attempts rather than relying on a narrow set of reasoning patterns. The consistent trend observed across representative and widely-used models supports the generality of our findings.  Inspired by this observation, we hypothesize that (1) a larger set of distinct answer candidates and (2) richer sequences of semantic transitions may afford more opportunities for discovering correct solution trajectories. This insight motivates us to propose DiScO, a framework that equips models with mechanisms to recognize their own thinking schemata and reinforces the generation of distinct transitions and solution paths during both training and inference.

\section{\textbf{Di}verse \textbf{Sc}hemata Policy \textbf{O}ptimization (DiScO)}

To enhance the reasoning ability of LRMs by explicitly modeling and encouraging diversity in their thinking schemata, we propose \textbf{Di}verse \textbf{Sc}hemata Policy \textbf{O}ptimization (DiScO), which consists of three key components: (1) {\em Schemata-Aware Supervised Fine-Tuning}, which equips the model to mark reasoning transitions and identify intermediate answer candidates, fostering self-awareness of its thought dynamics; (2) {\em Diversity-Oriented Reinforcement Learning}, which encourages distinct transitions and diverse solution trajectories through a tailored reward function; and (3) {\em Inference-Time Diversity Enhancement}, which introduces two simple yet effective strategies that dynamically optimize the reasoning chain at inference time. These components enable the model to think more ``diversely'', leading to more flexible reasoning.

\subsection{Schemata-Aware Supervised Fine-Tuning}

To equip the model with schemata awareness, i.e., enabling it to explicitly recognize and annotate the thinking schemata within its own reasoning chains, we introduce two specialized tokens. Specifically, \texttt{$\backslash$ReasoningTransition} signifies reasoning transitions, while \texttt{$\backslash$AnswerCandidate} denotes answer candidates. These tokens render the model’s thinking schemata observable and amenable to analysis without the need for additional annotation. To scale the construction of schemata-aware supervision data, we employ an LLM annotator to insert these markers into reasoning trajectories and then perform SFT on the annotated data.

\begin{figure*}[t]
\centering
\includegraphics[width=0.93\linewidth]{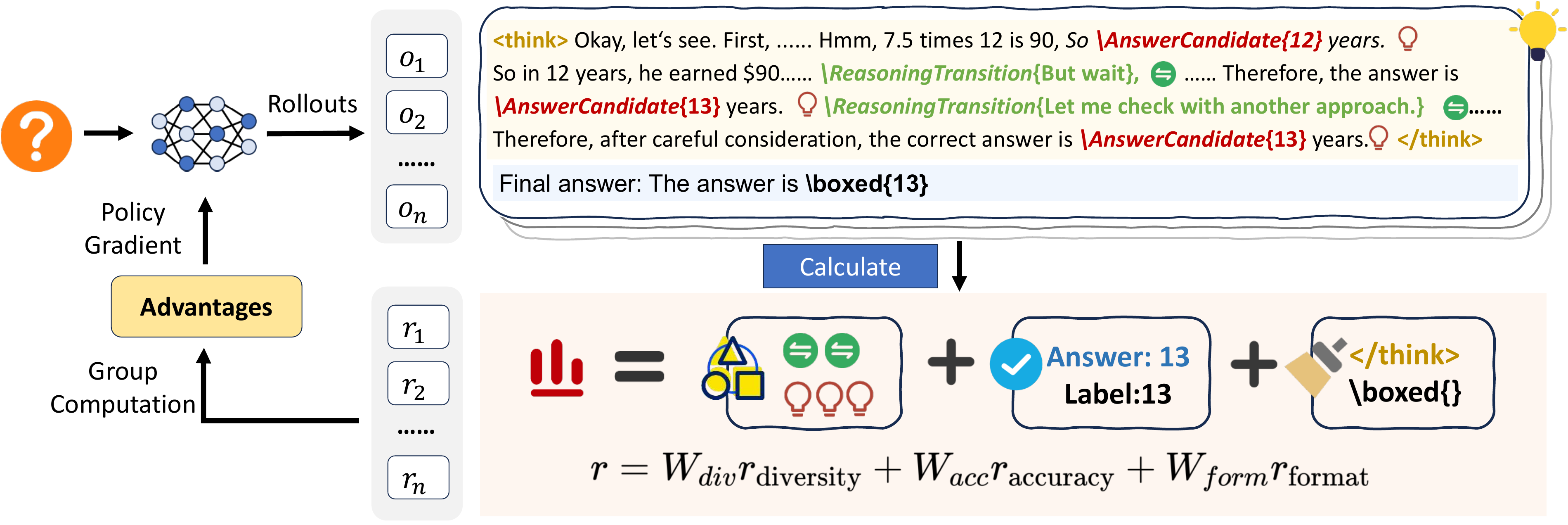} 
\caption{An illustration of the DiScO framework. During training, the model generates reasoning chains and receives rewards for the diversity of thinking schemata, structured formatting, and accurate final outputs.}
\label{fig:method}
\end{figure*}

\subsection{Diversity-Oriented Reinforcement Learning}

After equipping the model with schemata awareness, we employ reinforcement learning to encourage the generation of diverse thinking schemata during reasoning. The key insight is that models trained only with standard accuracy- or format-based rewards~\citep{shao2024deepseekmath, guo2025deepseek} tend to converge to narrow reasoning patterns, limiting their ability to explore alternative solution paths. To address this, we design a composite reward function that explicitly promotes diversity in both reasoning transitions and answer candidates, as follows:
\begin{equation}
    R = W_{\text{div}}R_{\text{div}} + W_{\text{acc}}R_{\text{acc}} + W_{\text{form}}R_{\text{form}},
\end{equation}
where $W_{*}$ denotes the corresponding weight assigned to each reward component.

\textbf{Diversity Reward.} To capture richness in reasoning trajectories, we measure four complementary aspects:
\begin{align}
\label{eq:div_reward}
R_{\text{div}} &= Cnt_A \cdot w_{cnt} + Cnt_T \cdot w_{cnt} \nonumber \\&+ Div_A \cdot w_{div} + Acc_A \cdot w_{true},\\
Cnt_A  
&= \min(N_{ans}, Cnt_A^{max}),
\label{eq:cnt_a} \\
Cnt_T  
&= \min(N_{thought}, Cnt_T^{max}),
\label{eq:cnt_t} \\
Div_A  
&= \min(N_{ans}^{uniq}, Div_A^{max}),
\label{eq:div_a} \\
Acc_A  
&= \min(N_{ans}^{true}, Acc_A^{max}).
\label{eq:acc_a}
\end{align}

Here $N_{ans}$ counts answer candidates, $N_{thought}$ counts reasoning transitions, $N_{ans}^{uniq}$ measures unique solution candidates, and $N_{ans}^{true}$ counts correct intermediate candidates. ${*}^{max}$ and $\omega_{*}$ are hyperparameters, where ${*}^{max}$ denotes the maximum truncation limit for the corresponding $N_{*}$, and $\omega_{*}$ represents its associated weight.

The above terms directly reflect the two core aspects of thinking schemata: $N_{thought}$ captures the richness of \textbf{Reasoning Transitions}, while $N_{ans}$ and its variants ($N_{ans}^{uniq}$, $N_{ans}^{true}$) operationalize the diversity and reliability of \textbf{Answer Candidates}. By rewarding both dimensions jointly, the model is encouraged to explore alternative inferential routes while maintaining plausible and accurate solution hypotheses. Further analyses (see Appendix~\ref{app:reward_semantic_validity}) demonstrate the reward resists marker-repetition hacking and rewards substantive reasoning shifts rather than surface paraphrase.

\textbf{Accuracy Reward.} The model is rewarded for producing a correct final answer:
\begin{equation}
    R_{\text{acc}} = 
    \begin{cases}
      1 & \text{if the final answer is correct},\\
      0 & \text{otherwise}.
    \end{cases}
\end{equation}

\textbf{Format Reward.} To ensure well-structured reasoning outputs (using the  \texttt{</think>}  tag to mark the end of thought and the \texttt{\textbackslash boxed\{\}}  to represent the final answer), we provide a lightweight reward for following the annotation format:
\begin{align}
    R_{\text{form}}(o) &= 0.5 \cdot [\![\,\texttt{"</think>"}\in o\,]\!] \\
    & + 0.5 \cdot [\![\,\texttt{"\textbackslash boxed"}\in o\,]\!]
\end{align}

By combining these components, DiScO aligns reinforcement learning not only with accuracy but also with the promotion of diverse thinking schemata, bridging the empirical link between diversity and performance observed in our analysis.

We adopt Group Relative Policy Optimization (GRPO)~\citep{shao2024deepseekmath, guo2025deepseek} as our reinforcement learning backbone. GRPO is a variant of PPO~\citep{schulman2017proximal} that avoids training an additional value function by leveraging group-wise relative comparisons of rollouts.

\subsection{Inference-Time Diversity Enhancement}
To encourage more diverse reasoning during inference on complex tasks, we introduce two simple yet effective Inference-Time Diversity Enhancement(ITDE) strategies that dynamically optimize the generated reasoning chain at inference time. Both strategies are triggered when the model reaches the maximum generation length without producing a final answer. Thus, they do not affect generations that already terminate normally.

\paragraph{Initial Truncation.} When triggered, we discard the earliest 20\% of tokens in the current reasoning chain and keep the remaining 80\% as the continuing context. The model then continues autoregressive generation conditioned on this retained suffix. This strategy simulates a cognitive reset, giving the model more test-time ``thinking space'' to reframe the problem and potentially explore new reasoning trajectories.

\paragraph{Truncation with Repetition Elimination.} This strategy extends Initial Truncation by additionally removing repetitive content. Specifically, we define repetitive reasoning as the occurrence of the same 15-word phrase more than once. Upon detection, we truncate the reasoning chain at the first occurrence and allow the model to continue generation from that position. This method prevents the model from being trapped in local loops and encourages exploration of more diverse and meaningful \mbox{reasoning} paths.

Together, these techniques aim to expand the inference-time search space, mitigate redundancy, and promote the diversity of thinking schemata. 

\section{Experiments}
\subsection{Experiment Setup}
\label{sec:experimental_setup}

\paragraph{Datasets} 
We use DeepSeek-R1-Distill-Qwen-7B/32B as our base models and build our implementation on the VeRL framework~\citep{sheng2024verl}. For schemata-aware SFT, we sample 885 samples from the OpenR1-Math-220k dataset and annotate the thinking schema by the Qwen3-Max, with the annotation prompt provided in the Appendix~\ref{prompt:Labeling}. We use 840 samples for training and 45 samples for validation to prevent overfitting.
For the GRPO stage, We randomly sampled 8k instances from the DeepScaler~\citep{shi2025deepscaler} dataset with accuracy ranging from 20\% to 80\% to avoid cases that are overly simple or overly complex, reserving 0.5\% as the validation set.\footnote{We have verified the SFT and GRPO training data to ensure that there is no overlap with the test sets utilized in our experiments.}  We use 8 rollouts per prompt, and a max length of 32768 tokens. Other training hyperparameters can be found in Appendix~\ref{app:hyperparameters}.

Following previous work on reasoning evaluation~\citep{wang2025highentropy,liu-etal-2025-safe}, we select five mathematical reasoning benchmarks of varying difficulty levels. Specifically, we adopt GSM8K~\citep{cobbe2021gsm8k},  MATH500~\citep{hendrycks2024math} as standard benchmarks, along with the more challenging competition-level benchmarks including AIME 2024~\citep{li2024numinamath}, AIME 2025~\citep{li2024numinamath}, and AMC 2023~\citep{li2024numinamath}. Additionally, although our training focuses on math, we extend our evaluation to out-of-domain (OOD) tasks to assess the generalizability of our approach, including ARC-c~\citep{clark2018arc}(OpenDomain Reasoning), GPQA-diamond~\citep{rein2024gpqa}(Science Graduate Knowledge), MMLU-Pro~\citep{wang2024mmlupro}(Reasoning-focused Questions from Academic Exams and Textbooks)

\paragraph{Baselines} We leverage a diverse range of Frontier LLMs and Open-Source Reasoning LLMs as our baseline models. Frontier LLMs include leading proprietary models such as GPT-5~\citep{singh2025openai}, DeepSeek-V4~\citep{deepseekai2026deepseekv4}, which are recognized for their advanced reasoning capabilities. Open-Source Reasoning LLMs feature prominent models like Mathstral-7B~\citep{mistral2024mathstral} and NuminaMath-72B~\citep{li2024numinamath}, which are widely benchmarked in mathematical tasks. The 7B and 32B parameter-scaled cohorts include base and instruction-tuned variants such as Qwen2.5-Math-7B/32B, ReasonFlux-7B/32B~\citep{yang2025reasonflux}, and QwQ-32B~\citep{qwq-32b-preview}, which are evaluated to explore the trade-offs between model size and performance. In addition, we compare with four 7B-scale diversity-based RL methods, namely Entropy-RL~\citep{cui2025entropy}, GRPO w/Clip-higher~\citep{yu2025dapo}, Pass@k Training~\citep{chen2025passk}, and DIVER~\citep{hu2025diver}. To ensure a fair comparison, we keep the inference setup consistent across all baselines, using the same prompt in Appendix~\ref{prompt:Inference} and the same generation hyperparameters in Appendix~\ref{app:hyperparameters}.

\subsection{Results \& Analyses}

\begin{table*}[t!]
\centering
\resizebox{\textwidth}{!}{%
\begin{tabular}{lccccccccc}
\toprule
Model & \cellcolor{lightgreen}MATH-500 & \cellcolor{lightgreen}GSM8K & \cellcolor{lightred}AIME 2024 & \cellcolor{lightred}AIME 2025 & \cellcolor{lightred}AMC 2023 & \cellcolor{lightyellow}ARC-c & \cellcolor{lightyellow}GPQA* & \cellcolor{lightyellow}MMLU-Pro & \cellcolor{lightblue}Average \\
\midrule
\multicolumn{10}{c}{\bf Reference Models} \\
\midrule
\multicolumn{10}{l}{\underline{\textit{Closed-source frontier}}} \\
GPT-4o & \cellcolor{lightgreen}76.6 & \cellcolor{lightgreen}89.5 & \cellcolor{lightred}16.7 & \cellcolor{lightred}26.7 & \cellcolor{lightred}47.5 & \cellcolor{lightyellow}72.8 & \cellcolor{lightyellow}53.6 & \cellcolor{lightyellow}72.6 & \cellcolor{lightblue}57.0 \\
Claude3.5-Sonnet & \cellcolor{lightgreen}78.3 & \cellcolor{lightgreen}96.4 & \cellcolor{lightred}16.0 & \cellcolor{lightred}43.3 & \cellcolor{lightred}80.0 & \cellcolor{lightyellow}76.7 & \cellcolor{lightyellow}59.4 & \cellcolor{lightyellow}75.1 & \cellcolor{lightblue}65.7 \\
GPT-o1-mini & \cellcolor{lightgreen}90.0 & \cellcolor{lightgreen}95.8 & \cellcolor{lightred}56.7 & \cellcolor{lightred}50.8 & \cellcolor{lightred}85.0 & \cellcolor{lightyellow}90.4 & \cellcolor{lightyellow}60.0 & \cellcolor{lightyellow}80.3 & \cellcolor{lightblue}72.8 \\
GPT-o1-preview & \cellcolor{lightgreen}85.5 & \cellcolor{lightgreen}94.9 & \cellcolor{lightred}44.6 & \cellcolor{lightred}46.7 & \cellcolor{lightred}85.0 & \cellcolor{lightyellow}86.8 & \cellcolor{lightyellow}73.3 & \cellcolor{lightyellow}77.8 & \cellcolor{lightblue}74.4 \\
Qwen3-max & \cellcolor{lightgreen}95.0 & \cellcolor{lightgreen}98.2 & \cellcolor{lightred}93.3 & \cellcolor{lightred}80.0 & \cellcolor{lightred}100.0 & \cellcolor{lightyellow}93.9 & \cellcolor{lightyellow}85.0 & \cellcolor{lightyellow}84.7 & \cellcolor{lightblue}91.3 \\
GPT-5(high) & \cellcolor{lightgreen}98.0 & \cellcolor{lightgreen}99.2 & \cellcolor{lightred}96.7 & \cellcolor{lightred}93.3 & \cellcolor{lightred}97.5 & \cellcolor{lightyellow}91.8 & \cellcolor{lightyellow}84.8 & \cellcolor{lightyellow}86.7 & \cellcolor{lightblue}93.8 \\
\multicolumn{10}{l}{\underline{\textit{Open-source frontier}}} \\ 
NuminaMath-72B-CoT & \cellcolor{lightgreen}64.0 & \cellcolor{lightgreen}91.4 & \cellcolor{lightred}3.3 & \cellcolor{lightred}13.3 & \cellcolor{lightred}70.0 & \cellcolor{lightyellow}73.9 & \cellcolor{lightyellow}34.4 & \cellcolor{lightyellow}40.9 & \cellcolor{lightblue}48.9 \\
LLaMA3.1-70B-Instruct & \cellcolor{lightgreen}65.4 & \cellcolor{lightgreen}91.7 & \cellcolor{lightred}16.7 & \cellcolor{lightred}3.3 & \cellcolor{lightred}50.0 & \cellcolor{lightyellow}85.2 & \cellcolor{lightyellow}47.6 & \cellcolor{lightyellow}66.3 & \cellcolor{lightblue}53.3 \\
DeepSeek-Coder-V2-Instruct & \cellcolor{lightgreen}75.3 & \cellcolor{lightgreen}94.9 & \cellcolor{lightred}13.3 & \cellcolor{lightred}26.7 & \cellcolor{lightred}57.5 & \cellcolor{lightyellow}82.5 & \cellcolor{lightyellow}43.0 & \cellcolor{lightyellow}48.0 & \cellcolor{lightblue}55.2 \\
LLaMA3.1-405B-Instruct & \cellcolor{lightgreen}73.8 & \cellcolor{lightgreen}96.4 & \cellcolor{lightred}23.3 & \cellcolor{lightred}10.0 & \cellcolor{lightred}50.0 & \cellcolor{lightyellow}85.6 & \cellcolor{lightyellow}48.2 & \cellcolor{lightyellow}73.0 & \cellcolor{lightblue}57.5 \\
Qwen2.5-Math-72B-Instruct & \cellcolor{lightgreen}85.6 & \cellcolor{lightgreen}95.5 & \cellcolor{lightred}30.0 & \cellcolor{lightred}26.7 & \cellcolor{lightred}70.0 & \cellcolor{lightyellow}84.4 & \cellcolor{lightyellow}42.1 & \cellcolor{lightyellow}72.5 & \cellcolor{lightblue}63.4 \\
DeepSeek-V3 & \cellcolor{lightgreen}90.2 & \cellcolor{lightgreen}94.2 & \cellcolor{lightred}47.7 & \cellcolor{lightred}39.2 & \cellcolor{lightred}80.0 & \cellcolor{lightyellow}84.7 & \cellcolor{lightyellow}57.8 & \cellcolor{lightyellow}74.0 & \cellcolor{lightblue}71.0 \\
DeepSeek-R1 & \cellcolor{lightgreen}91.2 & \cellcolor{lightgreen}86.7 & \cellcolor{lightred}100.0 & \cellcolor{lightred}90.0 & \cellcolor{lightred}64.1 & \cellcolor{lightyellow}89.6 & \cellcolor{lightyellow}71.2 & \cellcolor{lightyellow}83.9 & \cellcolor{lightblue}84.6 \\
DeepSeek-V4-pro & \cellcolor{lightgreen}94.8 & \cellcolor{lightgreen}97.5 & \cellcolor{lightred}96.7 & \cellcolor{lightred}93.3 & \cellcolor{lightred}100.0 & \cellcolor{lightyellow}95.5 & \cellcolor{lightyellow}90.1 & \cellcolor{lightyellow}87.5 & \cellcolor{lightblue}94.4 \\

\midrule
\multicolumn{10}{c}{\bf 7B-Scale} \\
\midrule
\multicolumn{10}{l}{\underline{\textit{General instruction-tuned models}}} \\
LLaMA3.1-8B-Instruct & \cellcolor{lightgreen}51.4 & \cellcolor{lightgreen}82.4 & \cellcolor{lightred}6.7 & \cellcolor{lightred}0.0 & \cellcolor{lightred}25.0 & \cellcolor{lightyellow}22.7 & \cellcolor{lightyellow}7.6 & \cellcolor{lightyellow}16.6 & \cellcolor{lightblue}26.6 \\
Mathstral-7B-v0.1 & \cellcolor{lightgreen}57.8 & \cellcolor{lightgreen}77.1 & \cellcolor{lightred}0.0 & \cellcolor{lightred}0.0 & \cellcolor{lightred}37.5 & \cellcolor{lightyellow}22.4 & \cellcolor{lightyellow}7.1 & \cellcolor{lightyellow}14.2 & \cellcolor{lightblue}27.0 \\
\multicolumn{10}{l}{\underline{\textit{Math instruction-tuned models}}} \\
Qwen2.5-Math-7B & \cellcolor{lightgreen}58.8 & \cellcolor{lightgreen}\textbf{95.0} & \cellcolor{lightred}16.7 & \cellcolor{lightred}3.3 & \cellcolor{lightred}22.5 & \cellcolor{lightyellow}65.3 & \cellcolor{lightyellow}27.7 & \cellcolor{lightyellow}45.5 & \cellcolor{lightblue}41.9 \\
Qwen2.5-Math-7B-Instruct & \cellcolor{lightgreen}82.6 & \cellcolor{lightgreen}\textbf{95.0} & \cellcolor{lightred}13.3 & \cellcolor{lightred}16.7 & \cellcolor{lightred}28.3 & \cellcolor{lightyellow}83.6 & \cellcolor{lightyellow}40.6 & \cellcolor{lightyellow}47.6 & \cellcolor{lightblue}51.0 \\
\multicolumn{10}{l}{\underline{\textit{Reasoning-specialized models}}} \\
SuperCorrect-7B & \cellcolor{lightgreen}70.2 & \cellcolor{lightgreen}84.7 & \cellcolor{lightred}26.7 & \cellcolor{lightred}13.3 & \cellcolor{lightred}37.5 & \cellcolor{lightyellow}67.3 & \cellcolor{lightyellow}20.7 & \cellcolor{lightyellow}33.7 & \cellcolor{lightblue}44.3 \\
DeepSeek-R1-Distill-Qwen-7B & \cellcolor{lightgreen}64.0 & \cellcolor{lightgreen}70.2 & \cellcolor{lightred}36.7 & \cellcolor{lightred}13.3 & \cellcolor{lightred}87.5 & \cellcolor{lightyellow}58.8 & \cellcolor{lightyellow}10.6 & \cellcolor{lightyellow}30.5 & \cellcolor{lightblue}46.5 \\
ReasonFlux-7B & \cellcolor{lightgreen}88.6 & \cellcolor{lightgreen}83.9 & \cellcolor{lightred}36.7 & \cellcolor{lightred}36.7 & \cellcolor{lightred}80.0 & \cellcolor{lightyellow}74.0 & \cellcolor{lightyellow}35.9 & \cellcolor{lightyellow}49.1 & \cellcolor{lightblue}60.6 \\
Qwen3-8B & \cellcolor{lightgreen}90.0 & \cellcolor{lightgreen}91.0 & \cellcolor{lightred}56.7 & \cellcolor{lightred}46.7 & \cellcolor{lightred}65.0 & \cellcolor{lightyellow}82.0 & \cellcolor{lightyellow}41.9 & \cellcolor{lightyellow}38.0 & \cellcolor{lightblue}63.9 \\
rStar-Math-7B & \cellcolor{lightgreen}89.4 & \cellcolor{lightgreen}95.0 & \cellcolor{lightred}50.0 & \cellcolor{lightred}46.7 & \cellcolor{lightred}87.5 & \cellcolor{lightyellow}79.3 & \cellcolor{lightyellow}41.8 & \cellcolor{lightyellow}50.3 & \cellcolor{lightblue}67.5 \\
\multicolumn{10}{l}{\underline{\textit{Diversity-oriented RL models}}} \\
Entropy-RL & \cellcolor{lightgreen}83.2 & \cellcolor{lightgreen}82.7 & \cellcolor{lightred}23.3 & \cellcolor{lightred}13.3 & \cellcolor{lightred}57.5 & \cellcolor{lightyellow}52.0 & \cellcolor{lightyellow}37.9 & \cellcolor{lightyellow}47.7 & \cellcolor{lightblue}49.7 \\
GRPO w/ Clip-higher & \cellcolor{lightgreen}80.8 & \cellcolor{lightgreen}82.4 & \cellcolor{lightred}20.0 & \cellcolor{lightred}16.7 & \cellcolor{lightred}57.5 & \cellcolor{lightyellow}81.6 & \cellcolor{lightyellow}35.2 & \cellcolor{lightyellow}46.7 & \cellcolor{lightblue}52.6 \\
Pass@k Training & \cellcolor{lightgreen}84.0 & \cellcolor{lightgreen}85.3 & \cellcolor{lightred}20.0 & \cellcolor{lightred}16.6 & \cellcolor{lightred}52.5 & \cellcolor{lightyellow}78.5 & \cellcolor{lightyellow}36.8 & \cellcolor{lightyellow}48.5 & \cellcolor{lightblue}52.8 \\
DIVER & \cellcolor{lightgreen}85.6 & \cellcolor{lightgreen}87.3 & \cellcolor{lightred}25.0 & \cellcolor{lightred}22.5 & \cellcolor{lightred}62.5 & \cellcolor{lightyellow}83.0 & \cellcolor{lightyellow}40.1 & \cellcolor{lightyellow}50.7 & \cellcolor{lightblue}57.1 \\
\multicolumn{10}{l}{\underline{\textit{Ours}}} \\
DiScO$_\text{w/o ITDE}$ & \cellcolor{lightgreen}94.8 & \cellcolor{lightgreen}92.3 & \cellcolor{lightred}66.7 & \cellcolor{lightred}\textbf{50.0} & \cellcolor{lightred}87.5 & \cellcolor{lightyellow}82.3 & \cellcolor{lightyellow}43.4 & \cellcolor{lightyellow}51.2 & \cellcolor{lightblue}71.0 \\
\textbf{DiScO} & \cellcolor{lightgreen}\textbf{95.6}\textsuperscript{**} & \cellcolor{lightgreen}93.7\textsuperscript{**} & \cellcolor{lightred}\textbf{70.0}\textsuperscript{**} & \cellcolor{lightred}\textbf{50.0}\textsuperscript{**} & \cellcolor{lightred}\textbf{92.5}\textsuperscript{**} & \cellcolor{lightyellow}\textbf{84.8}\textsuperscript{**} & \cellcolor{lightyellow}\textbf{45.0}\textsuperscript{**} & \cellcolor{lightyellow}\textbf{51.7}\textsuperscript{**} & \cellcolor{lightblue}\textbf{72.9}\textsuperscript{**} \\

\midrule
\multicolumn{10}{c}{\bf 32B-Scale} \\
\midrule
\multicolumn{10}{l}{\underline{\textit{Instruction-tuned models}}} \\
Qwen2.5-32B-Instruct & \cellcolor{lightgreen}79.4 & \cellcolor{lightgreen}94.4 & \cellcolor{lightred}16.5 & \cellcolor{lightred}13.3 & \cellcolor{lightred}64.0 & \cellcolor{lightyellow}85.5 & \cellcolor{lightyellow}48.8 & \cellcolor{lightyellow}71.0 & \cellcolor{lightblue}59.1 \\
\multicolumn{10}{l}{\underline{\textit{Reasoning-specialized models}}} \\
Sky-T1-32B-preview & \cellcolor{lightgreen}89.5 & \cellcolor{lightgreen}94.8 & \cellcolor{lightred}43.3 & \cellcolor{lightred}36.7 & \cellcolor{lightred}82.5 & \cellcolor{lightyellow}83.7 & \cellcolor{lightyellow}56.8 & \cellcolor{lightyellow}63.2 & \cellcolor{lightblue}68.8 \\
ReasonFlux-32B & \cellcolor{lightgreen}91.2 & \cellcolor{lightgreen}79.3 & \cellcolor{lightred}56.7 & \cellcolor{lightred}37.2 & \cellcolor{lightred}85.0 & \cellcolor{lightyellow}84.3 & \cellcolor{lightyellow}61.2 & \cellcolor{lightyellow}68.4 & \cellcolor{lightblue}70.4 \\
QwQ-32B-preview & \cellcolor{lightgreen}90.6 & \cellcolor{lightgreen}91.2 & \cellcolor{lightred}50.0 & \cellcolor{lightred}46.7 & \cellcolor{lightred}75.0 & \cellcolor{lightyellow}87.1 & \cellcolor{lightyellow}62.4 & \cellcolor{lightyellow}71.0 & \cellcolor{lightblue}71.8 \\
DeepSeek-R1-Distill-Qwen-32B & \cellcolor{lightgreen}93.4 & \cellcolor{lightgreen}94.0 & \cellcolor{lightred}80.0 & \cellcolor{lightred}60.0 & \cellcolor{lightred}\textbf{97.5} & \cellcolor{lightyellow}87.3 & \cellcolor{lightyellow}62.1 & \cellcolor{lightyellow}73.4 & \cellcolor{lightblue}80.9 \\
Qwen3-32B & \cellcolor{lightgreen}92.0 & \cellcolor{lightgreen}94.0 & \cellcolor{lightred}83.3 & \cellcolor{lightred}\textbf{73.3} & \cellcolor{lightred}92.5 & \cellcolor{lightyellow}92.0 & \cellcolor{lightyellow}62.1 & \cellcolor{lightyellow}71.0 & \cellcolor{lightblue}82.5\\
\multicolumn{10}{l}{\underline{\textit{Ours}}} \\
DiScO$_\text{w/o ITDE}$ & \cellcolor{lightgreen}91.4 & \cellcolor{lightgreen}94.5 & \cellcolor{lightred}\textbf{86.7} & \cellcolor{lightred}66.7 & \cellcolor{lightred}\textbf{97.5} & \cellcolor{lightyellow}86.4 & \cellcolor{lightyellow}60.5 & \cellcolor{lightyellow}73.2 & \cellcolor{lightblue}82.1 \\
\textbf{DiScO} & \cellcolor{lightgreen}\textbf{93.8} & \cellcolor{lightgreen}\textbf{96.4}\textsuperscript{**} & \cellcolor{lightred}\textbf{86.7}\textsuperscript{**} & \cellcolor{lightred}66.7\textsuperscript{**} & \cellcolor{lightred}\textbf{97.5} & \cellcolor{lightyellow}\textbf{88.0}\textsuperscript{*} & \cellcolor{lightyellow}\textbf{64.0}\textsuperscript{**} & \cellcolor{lightyellow}\textbf{74.5}\textsuperscript{**} & \cellcolor{lightblue}\textbf{83.5}\textsuperscript{**} \\
\bottomrule
\end{tabular}}
\caption{Pass@1 accuracy comparison on various benchmarks.
* denotes statistical significance at $p<0.05$, and ** denotes significance at $p<0.01$, comparing to the base model.
(\colorbox{lightgreen}{Standard}, \colorbox{lightred}{Hard}, \colorbox{lightyellow}{OOD}, and \colorbox{lightblue}{Average})
}
\label{tab:results}
\end{table*}

\begin{figure*}[t]
\centering
\includegraphics[width=0.9\linewidth]{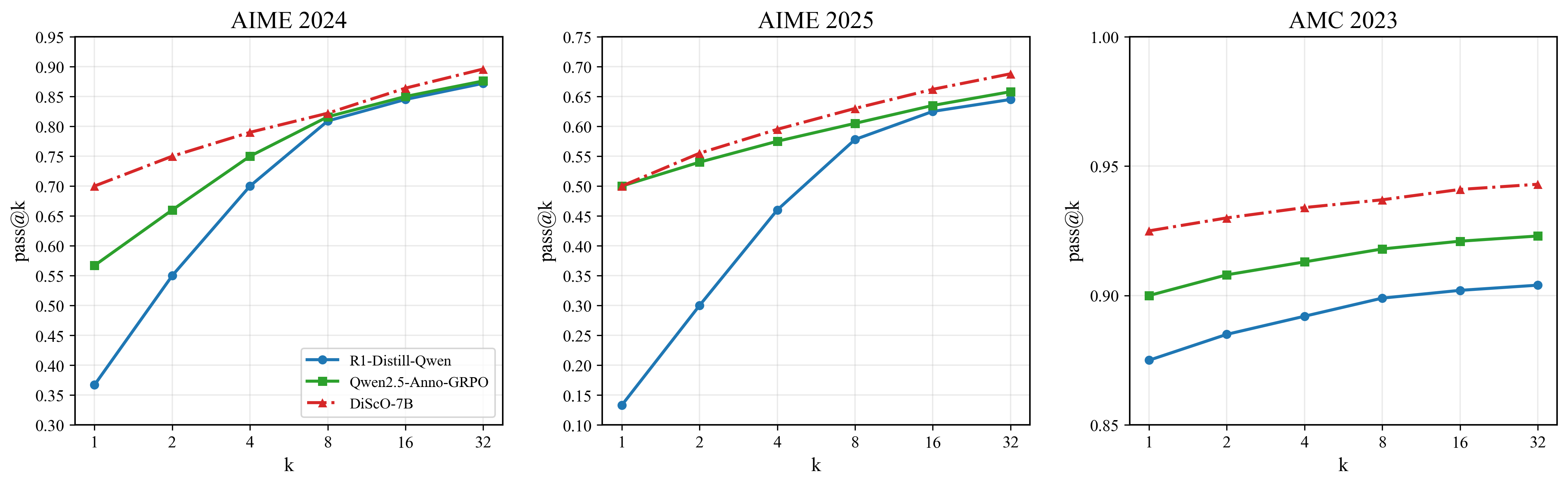} 
\caption{Pass@k curves of DiScO-7B compared with baselines across multiple benchmarks.}
\label{fig:passk_curves}
\end{figure*}


 \paragraph{DiScO Outperforms, Especially on Harder Tasks.}
To validate the overall effectiveness of DiScO, we compare it against both frontier LLMs and open-source reasoning models across reasoning benchmarks. As shown in Table~\ref{tab:results}, our proposed DiScO models achieve the best results at both scales. DiScO-7B obtains 95.6\% on MATH-500, 70.0\% on AIME 2024, and an average accuracy of 72.9\%, clearly surpassing other 7B models. DiScO-32B further improves to 93.8\% on MATH-500, 86.7\% on AIME 2024, 66.7\% on AIME 2025, and an average accuracy of 83.5\%, outperforming open-source baselines. 


We also observe that gains on near-saturated benchmarks such as MATH-500 and GSM8K are relatively modest, particularly for 32B models where baselines already exceed 93\% accuracy. We suspect this is due to ceiling effects that naturally limit observable improvements. However, on challenging competition-level benchmarks where substantial room for improvement remains, DiScO demonstrates more substantial gains. For instance, DiScO-32B achieves a 6.7\% absolute improvement over DeepSeek-R1-Distill-Qwen-32B on both AIME 2024 and AIME 2025. This pattern suggests that promoting diverse thinking schemata is especially valuable when models face problems that require extensive exploration of the reasoning space, with the benefits becoming increasingly apparent as task difficulty grows.

Component ablations in Appendix~\ref{app:ablation-study} further confirm that DiScO’s improvements come from its full schemata-aware optimization pipeline.


 \paragraph{Schemata-Level Diversity Beats Other Diversity Granularities.} Notably, DiScO also substantially outperforms recent diversity-oriented RL baseline. We attribute this gap to the granularity at which diversity is promoted. Token-level methods such as Clip-higher and Entropy-RL encourage localized variation through token-level optimization signals, which mainly improves micro-level exploration. Trajectory-level methods such as DIVER and Pass@k Training promote diversity across separate rollouts, but do not directly improve the internal structure of each reasoning chain. In contrast, DiScO promotes diversity at the schemata level by encouraging semantically meaningful reasoning transitions and answer candidates within a single trajectory. This enables the model to explore alternative solution strategies, revise intermediate hypotheses, and recover from errors more effectively, leading to particularly large gains on competition-level benchmarks.

\paragraph{DiScO Generalizes to Out-of-Domain Benchmarks.}
Although DiScO is trained exclusively on mathematical data, it consistently achieves the best performance among both 7B- and 32B-level models across all three OOD benchmarks. In particular, DiScO-32B outperforms the closest competitor QwQ-32B-preview on GPQA* and MMLU-Pro, with ARC-c showing a marginal trend($p=0.051$).

The OOD gains are smaller than the in-domain math gains, which is expected because the training signal is math-specific. Nevertheless, the consistent improvements suggest that schemata-level diversity promotes transferable
reasoning behaviors beyond the training domain, rather than merely improving math-specific pattern matching.

\paragraph{DiScO Improves Multi-Attempt Reasoning.} Beyond single-shot evaluation, we examine Pass@k accuracy to assess whether DiScO improves exploration across multiple reasoning attempts~\citep{yue2025does,chen2025pass}. Figure~\ref{fig:passk_curves} reports pass@k curves for DiScO-7B, the GRPO ablation, and the base model across three competition benchmarks. The GRPO ablation gradually converges toward the base model as $k$ grows, particularly on AIME 2024, suggesting that GRPO primarily reshapes the single-shot distribution rather than expanding the reasoning boundary. DiScO retains its advantage at large $k$, indicating genuine gains in reasoning capability.

\paragraph{Increased Diversity Enables Better Error Recovery.} 
To ascertain whether DiScO induces qualitatively distinct reasoning behavior, two human annotators annotated reasoning chains generated by DeepSeek-R1-Distill-Qwen and DiScO. From the annotated trajectories, we derive three metrics that characterize the model’s ability to revise or preserve intermediate hypotheses, as reported in Table~\ref{tab:error-recovery}. FWFC (First Wrong Final Correct) measures how often the model recovers from an erroneous initial attempt and ultimately produces the correct final answer. CorrLost measures how often a correct candidate appears in the reasoning trajectory but is later abandoned before the final answer. Recovery rate further normalizes FWFC by the frequency of initially incorrect attempts, thereby isolating the model's recovery capability from its baseline accuracy.
DiScO significantly improves FWFC across all benchmarks, with the most substantial gains on competition-level tasks. The recovery rate corroborates this by indicating genuine recovery capability rather than a higher rate of initiating incorrectly, as DiScO maintains a wide advantage when conditioned on erroneous initial attempts. CorrLost further demonstrates that DiScO abandons fewer correct intermediate answers than the baseline. These findings suggest that diverse thinking schemata enable models to reconsider early errors and converge on correct answers without destabilizing reasoning that is already on track.

\begin{table}[t!]
\centering
\resizebox{\columnwidth}{!}{%
\begin{tabular}{llccc}
\toprule
Benchmark & Model & FWFC $\uparrow$ & CorrLost $\downarrow$ & Recovery $\uparrow$ \\
\midrule
\multirow{2}{*}{MATH-50}
 & R1-Distill-Qwen & 6.0 & 18.0 & 21.4 \\
 & DiScO  & \textbf{42.0} & \textbf{0.0} & \textbf{87.5} \\
\midrule
\multirow{2}{*}{GSM8K-50}
 & R1-Distill-Qwen & 6.0 & 14.0 & 27.3 \\
 & DiScO  & \textbf{28.0} & \textbf{0.0} & \textbf{73.7} \\
\midrule
\multirow{2}{*}{AMC 2023}
 & R1-Distill-Qwen & 10.0 & 2.5 & 50.0 \\
 & DiScO  & \textbf{37.5} & \textbf{0.0} & \textbf{83.3} \\
\midrule
\multirow{2}{*}{AIME 2024}
 & R1-Distill-Qwen & 3.3 & 6.7 & 5.6 \\
 & DiScO  & \textbf{10.0} & \textbf{3.3} & \textbf{27.3} \\
\midrule
\multirow{2}{*}{AIME 2025}
 & R1-Distill-Qwen & 0.0 & 6.7 & 0.0 \\
 & DiScO  & \textbf{13.3} & \textbf{3.3} & \textbf{22.2} \\
\bottomrule
\end{tabular}}
\caption{Error recovery analysis on human-annotated trajectories (7B). MATH-50/GSM8K-50 indicate 50-problem subsets.}
\label{tab:error-recovery}
\end{table}

\section{Conclusion}

We introduce the concept of thinking schemata, characterized by reasoning transitions and answer candidates, as a novel perspective for understanding the reasoning process of large reasoning models. Through empirical analysis, we demonstrate that the diversity of thinking schemata correlates strongly with model performance, and we observe that existing models tend to follow narrow schemata, limiting their ability to explore alternative reasoning paths. Motivated by these findings, we propose DiScO, a framework that combines schemata-aware supervised fine-tuning, diversity-oriented reinforcement learning, and inference-time truncation strategies to systematically promote diverse reasoning during both training and inference. Extensive experiments demonstrate that DiScO consistently outperforms same-scale open-source baselines and several strong proprietary models, while narrowing the gap to frontier systems. Further analyses show that it enables models to explore more diverse solution paths and recover from suboptimal initial directions. We believe that encouraging diverse thinking schemata presents a promising direction for developing more human-like reasoning systems.

\section*{Limitations}
While DiScO consistently improves reasoning performance across model scales and benchmarks, a few limitations remain. The schemata-aware SFT data is annotated by an LLM and is therefore imperfect, which may bound the upper performance ceiling, and the annotation stage itself incurs non-trivial cost that motivates cheaper alternatives such as logits-based signals or probing over hidden states. Finally, our training and primary evaluation focus on mathematical reasoning, and although DiScO transfers to out-of-domain benchmarks with consistent gains, its effectiveness on substantially different reasoning regimes remains to be explored. We provide an extended discussion of these limitations, along with broader impacts and the relationship between diversity and scaling, in Appendix~\ref{app:discussion}.




\bibliography{custom}

\newpage
\appendix

\section{Detailed Related Works}

\paragraph{Large Reasoning Models} Recent advances in Large Reasoning Models, such as OpenAI o1~\citep{jaech2024openai-o1} and DeepSeek-R1~\citep{guo2025deepseek}, highlight the central role of reinforcement learning (RL) in enabling reasoning capabilities that pre-training alone cannot provide, with especially notable gains in mathematical reasoning~\citep{deepmath103k2025} and code generation~\citep{zhuo2024bigcodebench}. DeepSeek-R1, for instance, demonstrates that large-scale RL with structured accuracy or test-based rewards, implemented via Group Relative Policy Optimization (GRPO), can induce sophisticated reasoning behaviors even before downstream alignment~\citep{guo2025deepseek}. As RL becomes a standard mechanism for improving LRMs, recent work has increasingly emphasized not only correctness but also diversity in reasoning, inspired by~\citet{yao2025diversity}. Consequently, a growing line of research investigates RL methods that explicitly encourage diverse multi-step reasoning. Recently, some work focuses on improving GRPO via more effective reward design. \citet{zhang2025grpo} introduces enhancements to GRPO for mathematical reasoning, incorporating a length-dependent accuracy reward, explicit penalties for incorrect answers, and a difficulty-aware advantage reweighting strategy, collectively improving learning efficiency. Meanwhile, \citet{hu2020learning} study how to adaptively leverage a shaping reward by formulating its use as a bi-level optimization problem. Given GRPO’s widespread use and demonstrated effectiveness across LRM training pipelines, we also build on this framework in our method.

\paragraph{Reasoning Patterns in Large Reasoning Models} Researchers have investigated reasoning patterns in LRMs and their effects on solving math problems. \citet{minegishi2025topology} introduces ``Topology of Reasoning,'' where hidden-state clusters at each reasoning step form structures whose cycle frequency, graph diameter, and small‑world characteristics correlate with model capacity and task difficulty, offering interpretable graph-theoretic insights into why reasoning‑optimized LLMs perform better. \citet{tian2025think} presents a multi-round test-time thinking strategy where a model, given only its previous final answer (not its chain of thought), re-answers the same question across rounds, yielding consistent accuracy improvements without extra training. \citet{marjanovic2025deepseek} reveals that LRMs construct structured, multi-stage chains, starting with problem definition, followed by iterative “blooming” breakdowns and “reconstruction” reflections, before a final decision—showing that overly long reasoning degrades performance. \citet{an2025don} argues LRMs often bloat reasoning steps (i.e. “overthinking”) and proposes dynamically pruning inefficient sub-patterns in multi-stage chains, yielding more concise, resource-efficient, and accurate outcomes. \citep{yang2025reasonflux} enhances LRMs by guiding them through efficient multi-step reasoning with scalable, hierarchical thought templates, outperforming flat chain-of-thought methods. \citet{wen2025thinkpatterns} shows that while smaller LLMs benefit from structured “thinking'' patterns like decomposition, self‑ask, self‑debate, and self‑critic, larger models perform best with simpler, unstructured monologue-style reasoning. \citet{lee2025cot} proposes automatically extracting, clustering, and interpreting diverse chain-of-thought strategies from outputs to predict and steer LRMs toward more effective patterns. Our definition of schemata differs from, and is more general than, the aforementioned thinking patterns, as it does not rely on any manual attribution or classification of reasoning chains. Instead, we focus on the answer candidates and the transitions between reasoning trajectories, aiming to improve the performance of LRMs by encouraging the diverse generation of these thinking schemata.

\paragraph{Diversity in Large Reasoning Models} To improve the training performance of LRMs, recent studies have proposed methods to explored diversity in LRMs during training. \citet{yao2025diversity} investigate the importance of promoting diversity during RL training and introduce Potential@k, a metric quantifying an LLM’s reasoning potential after RL training. Their work demonstrates a strong correlation between solution diversity and performance. To leverage this, the method integrates a token-level diversity objective into R1-zero training, enhancing exploration while maintaining stability. \citet{wang2025highentropy} focus on local branching by identifying high-entropy tokens as key divergence points and optimizing RL updates around them. Token-level methods, however, operate on localized entropy signals and therefore encourage diversity only at the micro-level. In contrast, an emerging line of trajectory-level approaches seeks to promote diversity at the scale of entire reasoning paths. One prominent direction draws on GFlowNets, which sample structured objects, such as reasoning traces, with probability proportional to reward, thereby enabling diverse and globally consistent exploration. \citet{hu2023amortizing} use Sub-trajectory Balance to amortize posterior sampling in LLMs. \citet{yu2025flow} and \citet{nair2025flowoptions} formulate multi-step reasoning as flows on  Directed Acyclic Graphs(DAGs) to encourage varied reasoning paths under minimal supervision; and \citet{younsi2025accurate} trains a PRM from MCTS data and fine-tunes a step-level GFlowNet with Subtrajectory Balance and PRM-based multiplicative rewards to generate accurate, diverse trajectories. Despite these successes, trajectory-level GFlowNet methods typically rely on structural assumptions such as a DAG-based formulation of trajectories and the intrinsic flow-matching constraints required by the framework. These inductive biases guide exploration but can constrain the reasoning space, limiting the model’s ability to capture free-form, spontaneous reasoning patterns that do not align with the assumed structure. While token-level methods improve local branching behavior, their reliance on localized entropy signals restricts diversity to micro-level perturbations rather than capturing meaningful variations in reasoning strategy. GFlowNet-based approaches operate at the trajectory level but primarily optimize the probability of generating high-reward trajectories, achieving “reward-proportional” diversity rather than truly semantic diversity. In contrast, our method directly evaluates and shapes complete rollout trajectories, focusing on how they shift perspectives, explore alternative solution avenues, and form globally distinct reasoning paths. This trajectory-level view captures richer forms of exploratory reasoning and leads to genuine semantic diversity that better reflects the breadth of human-like problem-solving. In parallel, another line of work has explored improving diversity through test-time scaling, where multiple independent reasoning trajectories are generated and aggregated (e.g., self-consistency~\citep{wangself}). These methods sample diverse reasoning paths and select or aggregate answers based on consistency or learned criteria, leading to substantial gains in reasoning benchmarks. However, they primarily promote diversity across trajectories, treating each reasoning chain as an independent sample without modifying its internal reasoning dynamics. Subsequent work further strengthens this paradigm by explicitly learning to aggregate multiple sampled trajectories~\citep{qi2025learningreasonparallelsamples}. In contrast, DiScO focuses on diversity within a single trajectory, encouraging perspective shifts and intermediate hypothesis exploration during the reasoning process itself. This enables behaviors such as revisiting earlier steps and self-correction, which are not directly induced by independent parallel sampling.

\section{Hyperparameters}
\label{app:hyperparameters}
During the SFT stage, we adopt the LLaMA-Factory framework~\citep{llamafactory} with a context length of 32K, learning rate of 1e-5, and train each model for 3 epochs. For reinforcement learning, we employ the VeRL framework~\citep{sheng2024verl} with a 16K context length, 8 rollouts, and batch sizes of 256 for the 7B model and 512 for the 32B model. The 7B model is trained for 4 epochs (112 steps) on 8 nodes with 8$\times$H20 GPUs, requiring approximately 55 hours, while the 32B model is trained for 4 epochs (72 steps) on 16 nodes with 8$\times$H20 GPUs, taking about 36 hours. We use checkpoints at step 50 (7B) and step 40 (32B), where both models achieve peak accuracy. Hyperparameter details are summarized in Table~\ref{tab:hyperparams}. During inference, we adopt different decoding configurations for Pass@1 and Pass@k evaluation. For Pass@1, we apply greedy decoding with temperature 0.0 and top-p 0.2 via vLLM~\citep{vllm} to obtain deterministic and precise generations. For Pass@k, which requires multiple samples per prompt to assess exploration across attempts, we use stochastic sampling with temperature 0.6 and top-p 0.8. To ensure fair comparison, all models are evaluated under identical decoding configurations within each setting and share the same inference prompt. The prompts used for training and inference are demonstrated in Appendix~\ref{prompt:Training} and Appendix~\ref{prompt:Inference}, respectively.

\begin{center}
\begin{tabular}{l c}
\toprule
\textbf{Hyperparameter} & \textbf{Value} \\
\midrule
$\beta$ &  0.001\\
$\epsilon$ & 0.2 \\
$W_{acc}$       & 2.0 \\
$W_{div}$       & 1.0 \\
$W_{form}$      & 1.0 \\
$\omega_{cnt}$            & 0.1 \\
$\omega_{div}$        & 0.2 \\
$\omega_{true}$          & 0.3 \\
$Max_{ACnt}$       & 15 \\
$Max_{TCnt}$          & 20 \\
$Max_{ADiv}$      & 15 \\
$Max_{AAcc}$       & 15 \\
\bottomrule
\end{tabular}
\captionof{table}{Hyperparameters used in GRPO training.}
\label{tab:hyperparams}
\end{center}


The 12 hyperparameters in Table~\ref{tab:hyperparams} were not obtained through grid search or extensive task-specific optimization. Instead, we set them once using simple heuristics based on preliminary observations and then kept them fixed across all experiments. Specifically, the weights of the reward components follow standard practice in composite reward design, where final-answer accuracy is assigned the largest weight, while the diversity and format rewards provide auxiliary shaping signals. The truncation limits for answer candidates and reasoning transitions are selected by inspecting the typical ranges observed in a small number of reasoning chains sampled from the base model, and are capped at moderate upper bounds to reduce the risk of reward hacking through excessive marker generation. The sub-component weights are chosen to reflect the intuitive relative importance of candidate counting, candidate uniqueness, and candidate correctness. The RL-specific parameters, including $\beta$ and $\epsilon$, follow standard GRPO configurations from prior work. The fact that DiScO achieves consistent improvements across two model scales (7B and 32B), three model families (Qwen, LLaMA(Table~~\ref{tab:cross_model_family}), and comparisons with frontier models), and eight benchmarks (five mathematical and three out-of-domain) using the same set of heuristically chosen hyperparameters provides strong evidence for robustness. No per-dataset or per-scale tuning was performed.

\section{Semantic Validity of the Count-Based Diversity Reward}
\label{app:reward_semantic_validity}

A natural concern is whether the count-based diversity reward can be hacked by inserting transition markers that introduce only stylistic or lexical variation, without corresponding shifts in reasoning strategy (e.g., algebraic versus geometric approaches, computational versus formal derivation). We address this concern from two angles, first through the structural design of the reward itself, and then through an embedding-based semantic validation.

At the structural level, although the diversity reward uses count-based terms for efficiency, it is not a raw marker-counting objective. The uniqueness term $Div_A$ rewards only distinct answer candidates, while the truncation caps in Equations~\eqref{eq:cnt_a}--\eqref{eq:acc_a} prevent unbounded reward gains from excessive marker generation. Therefore, repetitive insertion of schemata markers cannot indefinitely increase the reward. As shown in Appendix~\ref{app:repeat-example}, cases with inflated numbers of reasoning transitions or answer candidates are mainly attributable to pathological repetition in the base model, further motivating these constraints in our reward design.

Sentence embeddings capture semantic content rather than surface form, so two transition segments that paraphrase the same reasoning will receive similar embeddings, while transitions that genuinely shift between different reasoning strategies will receive dissimilar embeddings. Under this interpretation, an embedding-based diversity score directly distinguishes paraphrastic restarts from substantive strategy shifts. We compute this score by measuring pairwise similarity between reasoning segments, and find a Spearman correlation of $\rho=0.57$ ($p<0.01$) with our count-based diversity reward. To further verify that the model-emitted markers correspond to genuine semantic shifts rather than stylistic insertions, we compare the embedding distance between segments immediately before and after each \texttt{\textbackslash ReasoningTransition} marker against the same distance computed at randomly sampled positions within the same trajectories. Model-chosen positions exhibit significantly larger before/after embedding distances than random positions ([0.58 vs. 0.72], $p<0.01$), confirming that the learned markers sit at locations of substantive reasoning shifts rather than arbitrary stylistic breaks. This indicates that count-based diversity tracks semantic-level diversity rather than marker frequency, and that the transitions DiScO learns to generate correspond to genuine shifts in reasoning content.

To further verify that the markers DiScO learns to emit correspond to genuine reasoning shifts rather than stylistic insertions, two annotators independently reviewed the transitions between \texttt{\textbackslash ReasoningTransition} markers from sampled 100 DiScO-7B's rollouts. For each marker, annotators judged whether the surrounding context represented a genuine shift in reasoning strategy (e.g., switching from numerical to algebraic manipulation, abandoning a prior hypothesis, or adopting a new solution avenue) versus a stylistic break with no substantive change in approach. Inter-annotator agreement was 0.56; disagreements were resolved by discussion. Across the sample, 83\% of markers were judged to correspond to genuine semantic shifts, indicating that the RL stage preserves the semantic quality of marker placement rather than introducing superficial insertions. The remaining 17\% include cases where the model re-attempts the same calculation with minor variations or restates a prior intermediate conclusion; these are not catastrophic for the framework, as the diversity reward's uniqueness term and truncation caps limit the reward gain from such repetition.

\section{Efficiency and Reasoning Length}
\label{appendix:token_overhead}

To quantify the efficiency implications of promoting diverse thinking schemata, we analyze the average output token length together with accuracy across multiple benchmarks at the 7B scale, as summarized in Table~\ref{tab:efficiency_7b}. 

Across all benchmarks, encouraging diversity in thinking schemata leads to a consistent increase in output length of approximately 7--14\%, while yielding substantial gains in reasoning accuracy, particularly on challenging datasets such as AIME where exploration of alternative reasoning trajectories is critical. This indicates a favorable cost--performance trade-off, as the additional tokens directly correspond to richer reasoning transitions and a larger set of answer candidates. In aggregate, DiScO introduces only a moderate increase of approximately 8--12\% in token count; since autoregressive inference latency scales approximately linearly with sequence length, the resulting latency overhead is similarly modest. On A100-80G GPUs with throughput of around 1.2k tokens per second for 7B models, this corresponds to less than one second of additional latency per query in typical settings, which we find to be well justified by the observed accuracy improvements. Compared to alternative strategies such as parallel sampling with aggregation (e.g., majority voting over $N$ independent reasoning chains), which incur an $N\times$ increase in inference cost, DiScO achieves stronger single-trajectory performance by promoting diversity within a single rollout, resulting in significantly better efficiency. 

We further examine whether increasing the number of answer candidates and reasoning transitions may distract the model and degrade reasoning quality. Empirically, our error recovery analysis (Table~\ref{tab:error-recovery}) shows the opposite effect: DiScO substantially improves FWFC, indicating that the model more effectively navigates the expanded reasoning space and recovers from incorrect intermediate steps, rather than being hindered by increased exploration. Additionally, we observe that the token overhead is smaller on simpler benchmarks such as GSM8K (approximately 7\%) and slightly larger on more challenging tasks, suggesting that DiScO adaptively allocates reasoning effort in proportion to problem difficulty. Nevertheless, the increased context length remains a limitation of our approach, particularly in deployment scenarios with strict latency or cost constraints, and developing more compute-efficient mechanisms for inducing diversity without increasing sequence length is an important direction for future work.

\begin{table*}[htb]
\centering
\begin{tabular}{l l c c}
\toprule
Model & Benchmark & Avg. Tokens & Accuracy \\
\midrule
R1-Distill & MATH-500 & 1090.7 & 64.0\% \\
DiScO & MATH-500 & 1245.1 (+14.2\%) & 95.6\% (+31.6\%) \\
\midrule
R1-Distill & GSM8K & 293.0 & 70.2\% \\
DiScO & GSM8K & 314.3 (+7.3\%) & 93.7\% (+23.5\%) \\
\midrule
R1-Distill & AIME 2024 & 4394.1 & 36.7\% \\
DiScO & AIME 2024 & 4857.9 (+10.6\%) & 70.0\% (+33.3\%) \\
\midrule
R1-Distill & AIME 2025 & 4893.1 & 13.3\% \\
DiScO & AIME 2025 & 5236.3 (+7.0\%) & 50.0\% (+36.7\%) \\
\midrule
R1-Distill & AMC 2023 & 2917.5 & 87.5\% \\
DiScO & AMC 2023 & 3152.6 (+8.1\%) & 92.5\% (+5.0\%) \\
\bottomrule
\end{tabular}
\caption{Average output token lengths and accuracy comparison at the 7B scale. Relative changes compared to the baseline are shown in parentheses.}
\label{tab:efficiency_7b}
\end{table*}

\section{More Results}

\subsection{Ablation Study}
\label{app:ablation-study}

\paragraph{Schemata-Aware SFT Makes Thinking Schemata Optimizable, while Diversity-Oriented RL Drives the Main Gains.} For ablation analysis, we first fine-tune DeepSeek-R1-Distill-Qwen-7B/32B on schemata-annotated reasoning traces to obtain the annotation-enhanced variant Qwen2.5-Anno. We then perform standard GRPO training without the diversity reward, obtaining Qwen2.5-Anno-GRPO. As shown in Table~\ref{tab:ablation-results}, schemata-aware SFT alone exhibits different effects across model scales. At the 7B scale, Qwen2.5-Anno improves the average accuracy from 54.4\% to 65.6\%, which may partially come from the additional supervised reasoning data introduced during SFT. However, at the 32B scale, Qwen2.5-Anno slightly decreases the average accuracy from 85.0\% to 83.8\%, suggesting that directly injecting annotation behavior may interfere with the original reasoning patterns of a stronger model when not followed by effective RL optimization. In contrast, the comparison between Qwen2.5-Anno-GRPO and DiScO shows that diversity-oriented RL is the main source of the final gains. DiScO improves over standard GRPO from 73.8\% to 80.4\% at the 7B scale, and more notably from 82.3\% to 88.2\% at the 32B scale. These results indicate that schemata-aware SFT mainly serves as a preparation stage that makes thinking schemata observable and optimizable, while the diversity-oriented reward is crucial for turning such annotations into stronger reasoning performance.

\begin{table*}[htb]
\centering
\resizebox{\textwidth}{!}{%
\begin{tabular}{lrrrrrr}
\toprule
Model & MATH-500   & GSM8K& AIME 2024 & AIME 2025 & AMC 2023 & Average \\
\midrule
\multicolumn{7}{c}{\bf 7B-Scale} \\
\midrule
DeepSeek-R1-Distill-Qwen & 64.0 & 70.2 & 36.7 & 13.3 & 87.5  & 54.4 \\
Qwen2.5-Anno & 72.6 & 78.7 & 56.7 & 40.0 & 80.0 & 65.6 \\
Qwen2.5-Anno-GRPO & 88.7 & 87.6 & 66.7 & 43.3 & 82.5   & 73.8 \\
DiScO & \textbf{95.6} & \textbf{93.7} & \textbf{70.0} & \textbf{50.0} & \textbf{92.5}  & \textbf{80.4} \\
\midrule
\multicolumn{7}{c}{\bf 32B-Scale} \\
\midrule
DeepSeek-R1-Distill-Qwen & 93.4 & 94.0 & 80.0 & 60.0 & \textbf{97.5}  & 85.0 \\
Qwen2.5-Anno & 92.2 & 95.3 & 73.3 & 63.3 & 95.0  & 83.8 \\
Qwen2.5-Anno-GRPO & 92.8 & 96.0 & 76.7 & 46.7 & \textbf{97.5}  & 82.3 \\
Disco & \textbf{93.8} & \textbf{96.4} & \textbf{86.7} & \textbf{66.7} & \textbf{97.5} & \textbf{88.2} \\
\bottomrule
\end{tabular}}
\caption{Pass@1 accuracy comparison on various mathematical reasoning benchmarks.}
\label{tab:ablation-results}
\end{table*}

\paragraph{Additional SFT Data Helps, but Explicit Schemata Annotation Enables Stronger RL Optimization.} To disentangle the effect of additional supervised data from that of schemata annotation, we further provide a full ablation ladder with non-annotated SFT data in Table~\ref{tab:ablation-ladder}. For the 7B model, SFT without annotation improves the average accuracy from 54.4\% to 65.2\%, which is close to the 65.6\% achieved by Qwen2.5-Anno in Table~\ref{tab:ablation-results}. This suggests that, for weaker models, part of the SFT-stage improvement comes from exposure to additional supervised reasoning data. For the 32B model, SFT without annotation slightly improves the average accuracy from 85.0\% to 85.7\%, whereas schemata-aware SFT alone decreases it to 83.8\%. This supports the suspicion that annotation learning alone may disturb the reasoning behavior of already strong models. However, the full DiScO pipeline consistently outperforms the non-annotated SFT+GRPO variant at both scales, improving the average accuracy from 73.0\% to 80.4\% at the 7B scale and from 85.9\% to 88.2\% at the 32B scale. These results show that simply introducing additional SFT data and standard GRPO can improve performance, but the gains remain limited compared with diversity-oriented RL over explicit thinking schemata. In other words, non-annotated SFT provides useful reasoning supervision, while schemata annotation makes the reasoning process structurally accessible to the diversity reward, leading to stronger and more consistent improvements.

\begin{table*}[htb]
\centering
\resizebox{\textwidth}{!}{%
\begin{tabular}{lcccccc}
\toprule
Model & MATH-500 & GSM8K & AIME 2024 & AIME 2025 & AMC 2023 & Avg \\
\midrule
\multicolumn{7}{c}{\bf 7B-Scale} \\
\midrule
DeepSeek-R1-Distill-Qwen & 64.0 & 70.2 & 36.7 & 13.3 & 87.5  & 54.4 \\
+ SFT (w/o annotation) & 78.3 & 83.7 & 50.0 & 36.7 & 77.5  & 65.2 \\
+ SFT (w/o annotation) + GRPO & 83.6 & 85.7 & 63.3 & \textbf{50.0} & 82.5 & 73.0 \\
DiScO & \textbf{95.6} & \textbf{93.7} & \textbf{70.0} & \textbf{50.0} & \textbf{92.5}  & \textbf{80.4} \\
\midrule
\multicolumn{7}{c}{\bf 32B-Scale} \\
\midrule
DeepSeek-R1-Distill-Qwen & 93.4 & 94.0 & 80.0 & 60.0 & \textbf{97.5}  & 85.0 \\
+ SFT (w/o annotation) & 92.6 & 94.2 & 83.3 & 63.3 & 95.0 & 85.7 \\
+ SFT (w/o annotation) + GRPO & 92.8 & 95.3 & 83.3 & 63.3 & 95.0 & 85.9 \\
Disco & \textbf{93.8} & \textbf{96.4} & \textbf{86.7} & \textbf{66.7} & \textbf{97.5} & \textbf{88.2} \\
\bottomrule
\end{tabular}
}
\caption{Pass@1 accuracy comparison on various mathematical reasoning benchmarks.}
\label{tab:ablation-ladder}
\end{table*}

\paragraph{Both Diversity Reward Components Contribute Complementarily.}
To explore the individual contribution of each diversity reward component, we ablate the Answer Candidate reward and Reasoning Transition reward separately. For this ablation study, we fine-tune DeepSeek-R1-Distill-Qwen-7B with standard SFT, which serves as the baseline for comparison with Qwen2.5-Anno that has annotation capability. We further perform GRPO training without the diversity reward on Qwen2.5-Anno, obtaining Qwen2.5-Anno-GRPO. Table~\ref{tab:reward-ablation} presents results across five mathematical reasoning benchmarks.

We observe that both dimensions of schemata diversity contribute substantially to performance. As shown in Table~\ref{tab:reward-ablation}, removing either reward leads to consistent drops in average accuracy. Specifically, removing the Answer Candidate reward decreases performance from 80.4\% to 77.3\%, while removing the Reasoning Transition reward reduces it to 77.9\%. This indicates that the two objectives capture complementary aspects of reasoning diversity. The Answer Candidate reward encourages exploration of multiple solution hypotheses, whereas the Reasoning Transition reward fosters richer transitions between intermediate reasoning steps.

\begin{table*}[t]
\centering
\resizebox{\textwidth}{!}{%
\begin{tabular}{lrrrrrl}
\toprule
Model
& MATH-500 & GSM8K & AIME 2024 & AIME 2025 & AMC 2023 & Average \\
\midrule
DiScO & \textbf{95.6} & \textbf{93.7} & \textbf{70.0} & 50.0 & \textbf{92.5} & \textbf{80.4} \\
\quad w/o AnswerCandidate & 93.0  & 93.3 & 60.0 & 50.0 & 90.0 & 77.3 (-2.9)\\
\quad w/o ReasoningTransition & 93.6 & 93.3 & 56.7 & \textbf{53.3} & \textbf{92.5}  & 77.9 (-2.3)\\ 
Qwen2.5-Anno-GRPO & 88.7 & 87.6 & 66.7 & 43.3 & 82.5   & 73.8 (-6.6) \\
\bottomrule
\end{tabular}}
\caption{Ablation results of diverse reward designs, showing Pass@1 accuracy across benchmarks (7B-Scale).}
\label{tab:reward-ablation}
\end{table*}

\begin{figure*}[t]
\centering
\includegraphics[width=0.95\linewidth]{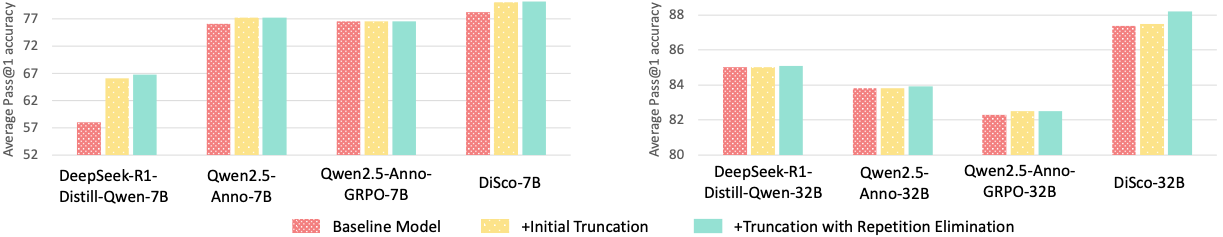} 
\caption{Average Pass@1 across mathematical reasoning benchmarks for inference-strategy ablations.}
\label{fig:bar-results}
\end{figure*}

\paragraph{Inference-Time Truncation Yields Stable Gains.}
To validate the effectiveness of our inference-time strategies, we evaluate two truncation methods across different models. As shown in Figure~\ref{fig:bar-results}, the truncation strategies consistently improve performance across different model scales. For example, DeepSeek-R1-Distill-Qwen-7B rises from 58.0\% to 66.8\% with repetition elimination, and DiScO-7B achieves the highest accuracy when combined with repetition elimination. At the 32B scale, DiScO-32B reaches 88.2\% with repetition elimination, establishing the strongest performance across all models. The largest benefits are observed on challenging datasets such as AIME 2025. These findings confirm that our lightweight truncation methods reduce redundancy in reasoning and improve overall accuracy, particularly when combined with diversity-oriented training.

\subsection{Diversity-Performance Correlation Analysis}
\label{app:diversity-correlation}

\begin{figure*}[htb]
\centering
\includegraphics[width=\linewidth]{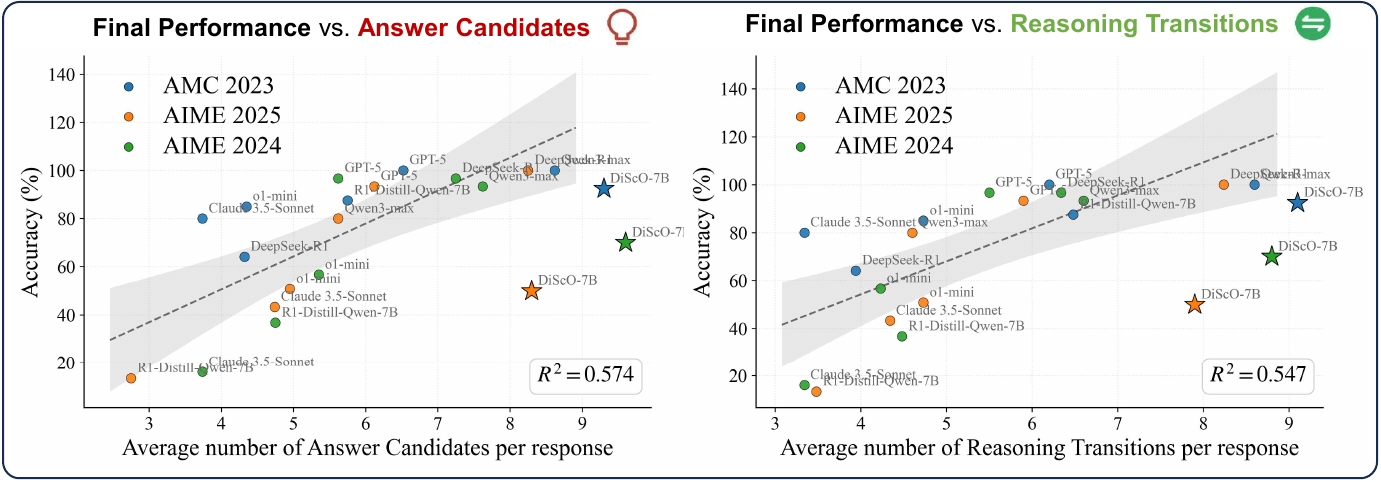} 
\caption{Enlarged scatter plots of accuracy versus answer candidates (left) and reasoning transitions (right), with regression fits. DiScO is shown but excluded from the regression.}
\label{fig:diversity-scatter-enlarged}
\end{figure*}

To support the preliminary analysis in Section~\ref{sec:verification}, we provide additional details on the experimental setup. We annotate reasoning chains per question from o1-mini~\citep{jaech2024openai-o1}, DeepSeek-R1-Distill-Qwen-7B~\citep{deepseekai2025deepseekr1}, Claude 3.5-Sonnet~\citep{anthropic2025claude35sonnet}, Qwen-math-plus, and Qwen3 on the AMC 2023, AIME 2024, and AIME 2025 benchmarks. Each chain is annotated by human annotators following the labeling protocol described in Appendix~\ref{app:human-study}, yielding per-chain counts of answer candidates and reasoning transitions, which are averaged within each (model, benchmark) pair to obtain summary statistics. For each diversity metric, we fit a linear regression between model accuracy and the metric across all (model, benchmark) data points. For completeness, we also annotate DiScO and visualize it in Figure~\ref{fig:intro}. However, DiScO is excluded from the correlation calculation itself, since the analysis is intended to motivate the design of DiScO rather than to validate it, and including DiScO would risk circularity. Enlarged versions of the scatter plots, with full model labels and per-benchmark regression fits, are shown in Figure~\ref{fig:diversity-scatter-enlarged}.

Although DiScO substantially increases schemata diversity over its base model, it does not sit at the extreme top-right of Figure~\ref{fig:intro}, where the largest frontier models reside. This observation should be interpreted alongside the broader pattern identified in Section~\ref{sec:verification}, where stronger base capability alone does not guarantee top performance, as evidenced by Claude 3.5-Sonnet and GPT-4o exhibiting limited schemata diversity and failing to consistently outperform smaller but more diverse counterparts. Taken together, these observations suggest that diversity and underlying capability are complementary contributors to reasoning performance, both necessary for top-tier results. First, without diverse thinking schemata, even high-capability models underperform. Second, without sufficient base capability, the inferential space opened by diverse transitions cannot be fully exploited, since each reasoning path must still be executed competently. DiScO directly addresses the first issue by explicitly training for diversity, while the second shapes the ceiling it can reach at a given parameter scale. From this perspective, DiScO does not substitute for capability scaling but supplies the diversity component that frontier-scale models obtain implicitly through massive pretraining and post-training compute. The consistent gains DiScO achieves over open-source baselines at the same parameter scale (Table~\ref{tab:results}) demonstrate the practical value of this axis, particularly for open-source settings where scaling parameters indefinitely is constrained.

\subsection{Generalization Analysis}
\label{app:cross_model_family}
To assess whether DiScO generalizes beyond the setting used in our main experiments, we further evaluate it on DeepSeek-R1-Distill-Llama-8B and Qwen3-8B. These two models allow us to examine two complementary questions: whether schemata annotations produced by a Qwen-based annotator can transfer to a different model family, and whether DiScO can also improve a native reasoning model rather than only reasoning-distilled models. Table~\ref{tab:cross_model_family} reports Pass@1 accuracy across five mathematical reasoning benchmarks.

\begin{table*}[htb]
\centering
\begin{tabular}{lccccc}
\toprule
\textbf{Model} & \textbf{MATH-500} & \textbf{GSM8K} & \textbf{AIME 2024} & \textbf{AIME 2025} & \textbf{AMC 2023} \\
\midrule
Distill-Llama-8B & 72.8 & 82.9 & 36.7 & 33.3 & 75.0 \\
DiScO-Llama-8B  & 78.8 & 86.3 & 43.3 & 40.0 & 90.0 \\
\midrule
Gain             & +6.0 & +3.4 & +6.6 & +6.7 & +15.0 \\
\bottomrule
\midrule
Qwen3-8B        & 90.0 & 91.0 & 46.7 & 46.7 & 65.0 \\
DiScO-Qwen3-8B  & 96.3 & 95.8 & 53.3 & 50.0 & 77.5 \\
\midrule
Gain             & +6.3 & +4.8 & +6.6 & +3.3 & +12.5 \\
\bottomrule
\end{tabular}
\caption{Cross-model-family evaluation on DeepSeek-R1-Distill-Llama-8B and Qwen3-8B. Results are reported as Pass@1 accuracy.}
\label{tab:cross_model_family}
\end{table*}

\paragraph{Qwen-Based Schemata Annotations Transfer Across Model Families.} Although our schemata-aware annotations are produced by Qwen3-Max, DiScO-Llama-8B achieves consistent improvements over DeepSeek-R1-Distill-Llama-8B across all five benchmarks, with an average gain of 7.54 percentage points. This result suggests that the annotations do not merely encode Qwen-family-specific reasoning habits. Instead, they provide model-agnostic structural supervision for identifying reasoning transitions and answer candidates, which can be learned and exploited by a different model family without modifying either the annotation data or the training pipeline.

\paragraph{DiScO Also Benefits Native Base Models.} Our main experiments are conducted on DeepSeek-R1-Distill-Qwen models, which inherit reasoning behaviors through distillation. To further examine whether DiScO can improve models with their own native reasoning ability, we apply the same pipeline to Qwen3-8B. As shown in Table~\ref{tab:cross_model_family}, DiScO-Qwen3-8B improves over Qwen3-8B on all five benchmarks. These results indicate that DiScO is not limited to enhancing reasoning-distilled models. Instead, explicitly optimizing diverse thinking schemata can further strengthen the reasoning behavior of models that already possess strong native reasoning capabilities.

\subsection{Pass@k Evaluation}
We focus on challenging benchmarks with lower Pass@1 accuracy, where the benefit of diverse exploration is more pronounced. As shown in Table~\ref{tab:pass-K-results}, DiScO surpasses all baselines on all three benchmarks suggesting that it improves reasoning not only within trajectories through richer transitions and answer candidates, but also across trajectories through more varied solution paths. This dual improvement indicates fundamental gains~\citep{cheng2025reasoning,zhu2025surprising} in reasoning capability rather than merely boosting single-shot performance.

\begin{table}[htb]
\centering
\resizebox{\linewidth}{!}{%
\begin{tabular}{lccc}
\toprule
Model & AIME 24 & AIME 25 & AMC 23 \\
\midrule
Qwen2.5-Math & 41.9 & 19.8 & 90.8 \\
Qwen2.5-Math-Inst & 24.8 & 30.2 & 84.0 \\
ReasonFlux & 74.3 & 60.8 & 93.1 \\
R1-Distill-Qwen & 80.9 & 57.8 & 89.9 \\
DiScO & \textbf{82.2} & \textbf{63.0} & \textbf{93.7} \\
\bottomrule
\end{tabular}
}
\caption{Pass@8(n=32) accuracy comparison on math benchmarks. All models are at the 7B parameter scale.}
\label{tab:pass-K-results}
\end{table}



\section{Prompt Design}
\label{app:prompt-design}

\subsection{Inference prompt design}
The adopted prompt for inference is shown below.

\label{prompt:Inference}
\begin{tcolorbox}[
    colback=blue!5!white, 
    colframe=blue!75!black, 
    title=Inference prompt design, 
    fonttitle=\bfseries, 
    boxrule=0.5mm, 
    arc=2mm, 
    left=2mm, 
    right=2mm, 
    top=2mm, 
    bottom=2mm 
]

Try to solve the following question step by step. If the final answer is obtained, use \textbackslash boxed\{\} to represent it.\\
\#\#\# Question:\{question\}\\
\#\#\# Assistant:
\end{tcolorbox}

\subsection{Training prompt design}
The adopted prompt for SFT and GRPO training is shown in Figure~\ref{fig:training-prompt}. 
\label{prompt:Training}
\begin{figure*}[h] 
\begin{tcolorbox}[
    colback=blue!5!white,
    colframe=blue!75!black,
    title=Training prompt design,
    fonttitle=\bfseries,
    boxrule=0.5mm,
    arc=2mm,
    left=2mm,
    right=2mm,
    top=2mm,
    bottom=2mm
]
Try to solve the following question step by step. Please show your reasoning chain according to the following rules:\\
1. First thinks about the reasoning chain in the mind and then provides the user with the answer. The reasoning chain is enclosed within \verb| | tags, i.e., \verb|| reasoning chain here \verb||. If the final answer is obtained, use \textbackslash boxed\{\} to represent it.\\
2. Label all segments that are potentially final results in the reasoning chain with \textbackslash AnswerCandidate{} format. DO NOT label all the possible intermediate results, ONLY label the ones that could be the final answers, no matter it's correct or wrong. Label as many as you could.\\
3. An example of the \textbackslash AnswerCandidate{} annotation: ``Wait, 5 times 360 is 1800, and 1800 divided by 36. Let's do that division: 1800 ÷ 36. Hmm, 36 times 50 is 1800, right? Because 36 x 50 is 1800. So, 1800 ÷ 36 = \textbackslash AnswerCandidate\{50\}. Therefore, the degrees for cherry pie would be \textbackslash AnswerCandidate\{50\} degrees."\\
4. Label all segments that indicate a shift in reasoning within the text reasoning chain using the \textbackslash ReasoningTransition{} format. Label as many as you could.\\
5.  An example of the \textbackslash ReasoningTransition{} annotation: ``\textbackslash ReasoningTransition\{Wait, maybe\} I messed up the daily progress. \textbackslash ReasoningTransition\{Wait, hold on\}. If the original total time is T days, then when they switch to the new equipment after 1/3 of the tunnel is done, which took T/3 days, and then the remaining 2/3 is done at a slower daily rate." \\
\#\#\# Question:\{question\}\\
\#\#\# Assistant:
\end{tcolorbox}
\caption{Prompt for Training.}
\label{fig:training-prompt}
\end{figure*}

\subsection{Labeling prompt design}
As described in Section~\ref{sec:experimental_setup}, we construct schemata-aware SFT data by sampling question-reasoning-answer triplets from OpenR1-Math-220k and annotating the reasoning chains with Qwen-Max. The adopted labeling prompt is shown in Figure~\ref{fig:labeling-prompt}. In the code implementation, we use the two tags ``AnswerCandidate" and ``ReasoningTransition" to represent ``Answer Candidate" and ``Reasoning Transition" respectively.

To assess the quality of automated annotations produced by Qwen3, we conducted a human validation study on 50 randomly sampled reasoning chains from the OpenR1-Math-220k dataset. Two graduate students with backgrounds in mathematics independently reviewed each annotation and judged whether the labeled reasoning transitions and answer candidates were correct according to the definitions. The automated annotations achieved F1-scores of 77.48\% for reasoning transitions and 84.87\% for answer candidates, respectively, with an inter-annotator Cohen's Kappa of 0.74 and 0.69 respectively, indicating substantial agreement and supporting the validity of our analysis.

\label{prompt:Labeling}
\begin{figure*}[htbp] 
\begin{tcolorbox}[
    colback=blue!5!white, 
    colframe=blue!75!black, 
    title=Labeling prompt design, 
    fonttitle=\bfseries, 
    boxrule=0.5mm, 
    arc=2mm, 
    left=2mm, 
    right=2mm, 
    top=2mm, 
    bottom=2mm 
]

The following is the reasoning chain that is used to answer a difficult math problem. Please process the reasoning chain according to the following rules:\\
1. Label all segments that are potentially final results in the reasoning chain with \textbackslash AnswerCandidate\{\} format. DO NOT label all the possible intermediate results, ONLY label the ones that could be the final answers, no matter it's correct or wrong. Label as many as you could.\\
2. An example of the \textbackslash AnswerCandidate\{\} annotation: ``Wait, 5 times 360 is 1800, and 1800 divided by 36. Let's do that division: 1800 ÷ 36. Hmm, 36 times 50 is 1800, right? Because 36 x 50 is 1800. So, 1800 ÷ 36 = \textbackslash AnswerCandidate\{50\}. Therefore, the degrees for cherry pie would be \textbackslash AnswerCandidate\{50\} degrees."\\
3. Label all segments that indicate a shift in reasoning within the text reasoning chain using the \textbackslash ReasoningTransition\{\} format. Label as many as you could.\\
4.  An example of the \textbackslash ReasoningTransition\{\} annotation: ``\textbackslash ReasoningTransition\{Wait, maybe\} I messed up the daily progress.\textbackslash ReasoningTransition\{Wait, hold on\}. If the original total time is T days, then when they switch to the new equipment after 1/3 of the tunnel is done, which took T/3 days, and then the remaining 2/3 is done at a slower daily rate"  \\
5. DO NOT change other parts and keep them exactly the same as the the original solution.\\

\#\#\# original solution: \\
\{solution\}\\
\#\#\# Result:\\
\end{tcolorbox}
\caption{Labeling prompt used for annotating candidate answers and reasoning transitions.}
\label{fig:labeling-prompt}
\end{figure*}

\section{Human Annotation Study Details}
\label{app:human-study}

\paragraph{Participants.} Two graduate students with backgrounds in mathematics were recruited as annotators for the human validation of automated schemata labels and for the manual review of model-generated reasoning chains. The study was approved by our institution's IRB, and informed consent was obtained from each participant prior to the task. Participants were informed that the task involved reading mathematical reasoning text only, that the data contained no personal or sensitive information, and that they could withdraw at any time without consequence.

\paragraph{Instructions provided to annotators.}
The instruction given to the annotators is shown in Figure~\ref{fig:instructions}, together with the formal definitions of \emph{Reasoning Transitions} and \emph{Answer Candidates}:

\begin{figure*}[htbp]
\begin{tcolorbox}[
    colback=blue!5!white, 
    colframe=blue!75!black, 
    title=Human labeling instructions, 
    fonttitle=\bfseries, 
    boxrule=0.5mm, 
    arc=2mm, 
    left=2mm, 
    right=2mm, 
    top=2mm, 
    bottom=2mm 
]

You will be shown a sequence of mathematical reasoning chains, each one
already annotated by an automatic labeling system. The system has marked
two kinds of spans:

\begin{itemize}
    \item \texttt{\textbackslash ReasoningTransition\{...\}}: a span of
    text that signals a shift in reasoning perspective or strategy
    (e.g., ``Wait,'', ``Let me try another approach,'', ``On the other
    hand,'').
    \item \texttt{\textbackslash AnswerCandidate\{...\}}: a value that
    the model proposes as a possible final answer to the problem
    (regardless of whether it turns out to be correct).
\end{itemize}

For each annotated span, please judge whether the label is correct
according to the definitions above. Mark each span as:

\begin{itemize}
    \item \textbf{Correct} -- the span genuinely matches the definition;
    \item \textbf{Incorrect} -- the span does not match the definition
    (false positive); or
    \item \textbf{Missing nearby} -- there is text close to this span
    that should have been labeled but was not (false negative).
\end{itemize}

Please base your judgement only on the definitions above, and not on
whether the model's final answer to the problem is correct. If you are
unsure about a span, mark it as ``Uncertain'' and add a brief note.

\end{tcolorbox}
\caption{Human labeling instructions.}
\label{fig:instructions}
\end{figure*}

\paragraph{Inter-annotator agreement.} The two annotators worked independently. Inter-annotator agreement was measured using Cohen's $\kappa$, yielding 0.74 for reasoning transitions and 0.69 for answer candidates, both indicating substantial agreement. Disagreements were resolved by discussion before computing the F1-scores of 77.48\% (reasoning transitions) and 84.87\% (answer candidates) reported.

\section{Annotation Cost and Model Size}
\label{app:annotation_cost}

The schemata-aware SFT stage requires annotated reasoning chains, which introduces an additional annotation cost before RL training. To better understand this cost--quality tradeoff, we evaluated annotators of different model sizes on the validation set. Table~\ref{tab:annotator_model_size} reports the F1 scores for identifying reasoning transitions and answer candidates.

\begin{table*}[htb]
\centering
\begin{tabular}{lcc}
\toprule
\textbf{Annotator Model} & 
\textbf{Reasoning Transition F1} & 
\textbf{Answer Candidate F1} \\
\midrule
Qwen3-8B   & 63.12 & 70.05 \\
Qwen3-32B  & 73.26 & 78.61 \\
Qwen3-max  & 77.48 & 84.87 \\
\bottomrule
\end{tabular}
\caption{Annotation quality across different annotator model sizes on the validation set.}
\label{tab:annotator_model_size}
\end{table*}

We ultimately use Qwen3-max because it provides the highest annotation quality. However, the F1 scores of 77.48\% for reasoning transitions and 84.87\% for answer candidates also indicate that the annotations are not
perfect. Despite this annotation noise, DiScO still achieves substantial improvements, suggesting that the framework is robust to imperfect schemata labels.

These results also suggest a practical path for reducing annotation cost. Although Qwen3-max gives the best labels, Qwen3-32B already achieves reasonable annotation quality, with 73.26\% F1 for ReasoningTransition
and 78.61\% F1 for AnswerCandidate. Therefore, practitioners may use a smaller annotator as a cost-effective alternative when annotation budget is limited. More broadly, as reasoning models become stronger, the schemata
labeling task can increasingly be performed by the model itself or by a smaller model from the same family. In this setting, the role of SFT is mainly to teach the model to follow the schemata-marking format, while the diversity-oriented RL stage is responsible for improving reasoning behavior.

\section{Case Study}

\subsection{Example of repetitive generation}
\label{app:repeat-example}
We observe that DeepSeek-R1-Distill-Qwen-7B tends to recycle its prior reasoning and generate repetitive content, especially on hard problems. An example is shown in Figure~\ref{fig:repetitive-example}

\begin{figure*}[htbp]
\begin{tcolorbox}[
    colback=white, 
    colframe=black, 
    title=Example: Response generated by DeepSeek-R1-Distill-Qwen-7B, 
    fonttitle=\bfseries, 
    boxrule=0.5mm, 
    arc=2mm, 
    left=2mm, 
    right=2mm, 
    top=2mm, 
    bottom=2mm 
]

Okay, so I have this problem where I need to express a double sum in terms of two known sums, p and q. Let me write down what I know.
......

But when I compute that, it's equal to \(\sum_{m=2}^{\infty}(1/m^2 - 1/m^3)\) = \(p - 1\) - \(q - 1\) = p - q. ......
Wait, maybe I need to consider that in the double sum, j and k are both starting from 1, so m starts from 2, but in the transformation, I have m - 1 terms, ......

\par\medskip

\textcolor{red!60!black}{But when I compute that, it's equal to \(\sum_{m=2}^{\infty}(1/m^2 - 1/m^3)\) = \(p - 1\) - \(q - 1\) = p - q. ......
Wait, maybe I need to consider that in the double sum, j and k are both starting from 1, so m starts from 2, but in the transformation, I have m - 1 terms, ......
}

......

\end{tcolorbox}
\caption{Example of repetitive generation.}
\label{fig:repetitive-example}
\end{figure*}

\subsection{Example of thinking schemata}
\begin{figure*}[hbt]
\centering
\includegraphics[width=\textwidth]{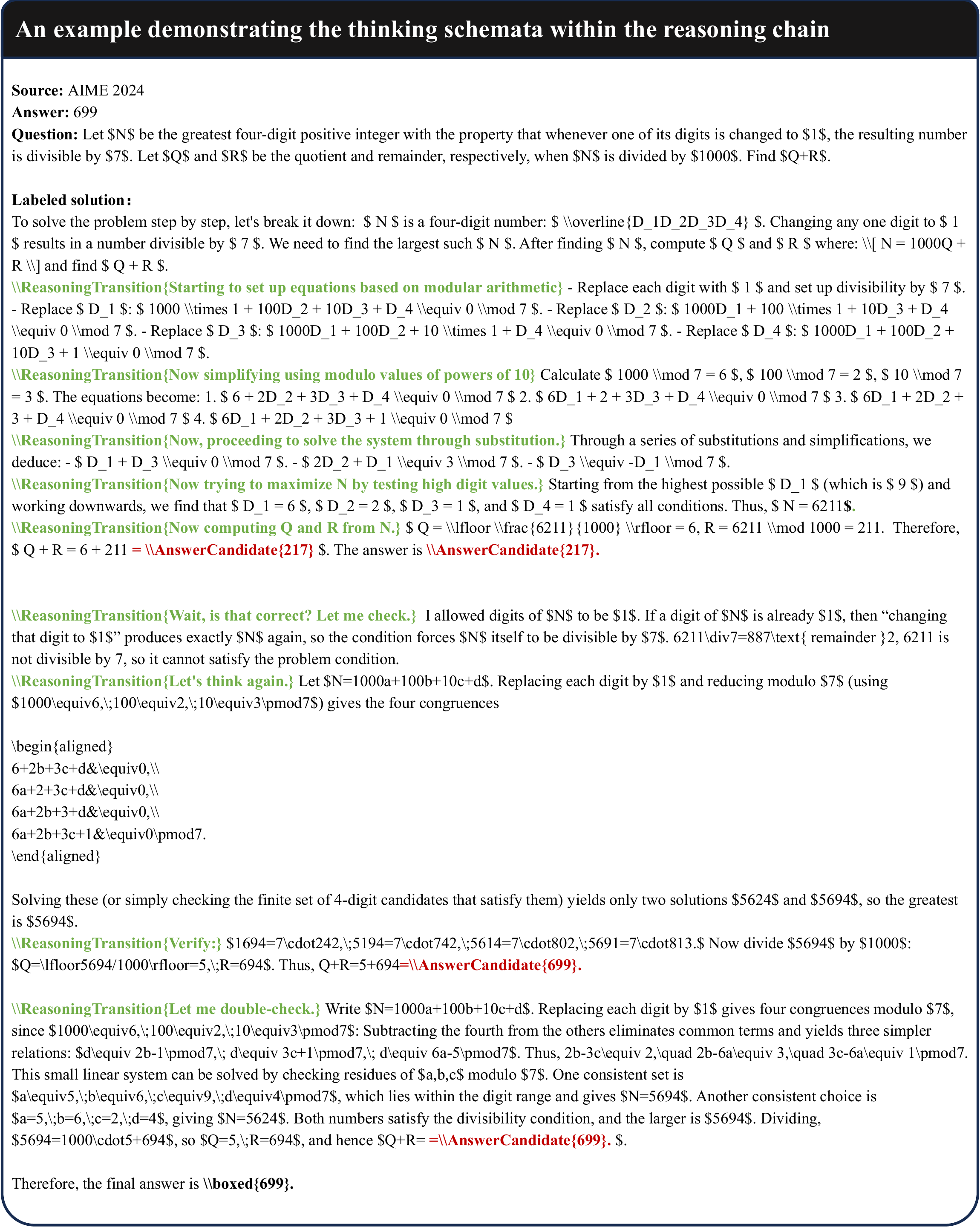} 
\caption{An example demonstrating the thinking schemata within a reasoning
chain. }
\label{fig:schemata-case}
\end{figure*}

Figure~\ref{fig:schemata-case} presents an example that demonstrates the thinking schemata within a reasoning
chain. 

Specifically, \textit{ReasoningTransition} reflects changes such as shifting from setting up equations to algebraic manipulation, or from a numerical computing to a formal derivation. It also characterizes cognitive processes, where the model engages in self-reflection, reassesses prior steps, and deliberately switches to an alternative reasoning strategy when the initial approach proves insufficient. \textit{AnswerCandidate} refers to the set of distinct final or intermediate solutions the model proposes during its reasoning process. 

\label{app:schemata-example}

\section{Discussion}
\label{app:discussion}

In this section, we reflect on the broader context of our work, including its limitations, potential societal impacts, the relationship between diversity and scaling, and directions for future research.

\subsection{Limitations}

While DiScO consistently improves reasoning performance across model scales and benchmarks, several limitations remain.

\paragraph{Inference-time overhead.} Encouraging diverse thinking schemata results in an approximately 10\% increase in average output token length (Appendix~\ref{appendix:token_overhead}), which translates into modest but non-negligible inference latency overhead. On A100-80G GPUs with throughput of around 1.2k tokens per second for 7B models, this corresponds to less than one second of additional latency per query in typical settings, which we find well justified by the observed accuracy improvements. Nevertheless, this overhead may be a concern in deployment scenarios with strict latency or cost constraints, and developing more compute-efficient mechanisms for inducing diversity without increasing sequence length is an important direction for future work.

\paragraph{Annotation reliability.} The schemata-aware SFT data is annotated by an LLM (Qwen-Max), which inevitably introduces noise. Human validation reports F1-scores of 77.48\% for reasoning transitions and 84.87\% for answer candidates, indicating that the labels are far from perfect, and this residual annotation noise may bound the upper performance ceiling of DiScO. Nevertheless, DiScO still achieves gains across model scales and benchmarks under these imperfect annotations, suggesting that the framework is robust to label noise and does not rely on pristine schemata supervision to be effective. The annotation stage itself also remains a non-trivial cost, since producing schemata labels with a strong LLM annotator scales linearly with the size of the SFT corpus and constitutes the dominant overhead beyond standard RL training. A promising direction for reducing this cost is to replace explicit LLM-based labeling with lightweight heuristic signals derived from the model's own internal states, such as logits-based indicators of perspective shifts (for example, entropy spikes or distributional discontinuities at transition tokens) or probing classifiers trained on a small annotated seed set to detect reasoning transitions and answer candidates directly from hidden representations. Such approaches could provide schemata supervision at a fraction of the current annotation cost, while complementary improvements in label quality through human-in-the-loop refinement or stronger labeling models remain worthwhile for further pushing the upper performance ceiling.

\paragraph{Reward design simplicity.} Our diversity reward uses count-based terms with truncation caps and uniqueness constraints. While we validate its semantic meaningfulness through an embedding-based diversity score (Appendix~\ref{app:reward_semantic_validity}), more sophisticated reward formulations, such as those based on learned semantic similarity or trajectory-level mutual information, may further improve performance at the cost of increased computational complexity.

\subsection{Broader Impacts}

Improving the reasoning capability and error-recovery behavior of large reasoning models has positive applications in education, scientific discovery, and any domain where multi-step reasoning support is valuable. The schemata-aware annotations also make the model's reasoning more interpretable and auditable. By explicitly marking transitions and answer candidates, DiScO produces reasoning traces that humans can more easily inspect and verify. This supports downstream applications such as automated tutoring systems, where surfacing alternative solution paths can help learners understand a problem from multiple angles, and human-AI collaboration, where transparent reasoning enables more effective oversight. However, stronger mathematical-reasoning models may be misused for automated cheating on educational assessments, particularly in unsupervised online testing scenarios. Additionally, the diversity-oriented reward could, in some cases, encourage the model to generate plausible-looking but incorrect intermediate answer candidates, which may mislead users who do not carefully verify the final answer.

\subsection{Diversity and Scaling}

To understand the broader implications of our findings, we discuss how diversity relates to scaling. Encouraging diverse thinking schemata through reinforcement learning requires several times more computation at the same parameter scale, but the improvements on challenging benchmarks justify this cost. Similar to recent findings that LRMs increasingly rely on compute-intensive post-training such as RL and reward modeling, we argue that diversity-based scaling is a worthwhile yet demanding direction. This suggests a complementary axis to the standard scaling laws based on parameters and data: rather than only scaling model size or pretraining tokens, scaling the diversity of the reasoning process itself may unlock further reasoning capability. This calls for both greater resources and more efficient optimization strategies, making it a crucial and challenging path for reasoning models.

\end{document}